\documentclass{article}

\PassOptionsToPackage{numbers, compress}{natbib}
\usepackage[final,main]{neurips_2026}

\usepackage[utf8]{inputenc}
\usepackage[T1]{fontenc}
\usepackage{hyperref}
\usepackage{url}
\usepackage{booktabs}
\usepackage{longtable}
\usepackage{amsmath}
\usepackage{amsfonts}
\usepackage{amssymb}
\usepackage{nicefrac}
\usepackage{microtype}
\usepackage[table]{xcolor}
\usepackage{graphicx}
\usepackage{bm}
\usepackage{caption}
\usepackage{subcaption}
\usepackage{placeins}
\usepackage[most]{tcolorbox}
\usepackage{enumitem}

\hypersetup{
    colorlinks=true,
    linkcolor=red,
    citecolor=teal,
    urlcolor=cyan,
}

\DeclareMathOperator*{\argmax}{arg\,max}

\newcommand{\R}{\mathbb{R}}
\newcommand{\E}{\mathbb{E}}
\newcommand{\Var}{\mathrm{Var}}
\newcommand{\method}{UtoMe}
\definecolor{tokenblue05}{HTML}{F6FBFF}
\definecolor{tokenblue10}{HTML}{ECF6FF}
\definecolor{tokenblue25}{HTML}{DDEEFF}
\definecolor{tokenblue50}{HTML}{C7E1FF}
\definecolor{utomeEbg}{HTML}{F1FFF4}
\definecolor{utomePbg}{HTML}{E2F7E9}
\definecolor{utomeSbg}{HTML}{D3EFDE}
\definecolor{referencegray}{HTML}{777777}
\definecolor{projectgreen}{HTML}{006400}
\definecolor{takeawayred}{HTML}{8E1B1B}
\definecolor{takeawaybg}{HTML}{FFF7F7}
\newcommand{\refval}[1]{\textcolor{referencegray}{#1}}
\newtcolorbox{takeawaybox}[1]{%
    colback=takeawaybg,
    colframe=takeawayred!65!black,
    title={\textbf{#1}},
    fonttitle=\bfseries,
    boxrule=0.8pt,
    arc=2pt,
    left=6pt,
    right=6pt,
    top=5pt,
    bottom=5pt}
\newcommand{\takeawaypoint}[2]{%
    \vspace{0.35em}
    \begin{takeawaybox}{#1}
        \small #2
    \end{takeawaybox}
    \vspace{0.35em}}

\title{UtoMe: Observation-Uncertainty-Guided Token Merging for Weather Foundation Models}

\author{%
    Yunbei Zhang\textsuperscript{1,2*\ensuremath{\dagger}}\quad
    Janet Wang\textsuperscript{1,2*}\quad
    Xi Xiao\textsuperscript{2}\quad
    Jihun Hamm\textsuperscript{1}\quad
    Xiao Wang\textsuperscript{2\ensuremath{\dagger}} \\
    \textsuperscript{1}Tulane University,\quad
    \textsuperscript{2}Oak Ridge National Laboratory
}

\begin{document}
\maketitle
\begingroup
\renewcommand{\thefootnote}{}
\footnotetext{\textsuperscript{*}Equal contribution.
\textsuperscript{\ensuremath{\dagger}}Corresponding authors.
This work was done during an internship at Oak Ridge National Laboratory.
Emails: \texttt{yzhang111@tulane.edu},
\texttt{wangx2@ornl.gov}.}
\endgroup

\begin{abstract}
Weather transformers process dense reanalysis fields whose reliability varies across variables, locations, and weather regimes. Token merging can reduce computation, but feature similarity alone does not account for this heterogeneous input uncertainty. We introduce \method, an observation-uncertainty-guided token merging framework that combines dynamics saliency with ensemble data assimilation spread as a relative input-reliability proxy. The resulting importance score protects selected tokens as anchors and calibrates bipartite matching between the remaining tokens. Three variants use external uncertainty at both training and inference, during training only, or neither stage. In ERA5 72-hour forecasting at two spatial resolutions, \method\ improves Overall error over train-time ToMe at matched token budgets. At $1.40625^\circ$ and $50\%$ token reduction, \method-E lowers Overall error from $67.32$ to $53.00$ at the same $0.55\times$ encoder FLOPs. Small reductions remain close to the full-token baseline; larger reductions incur measurable accuracy losses. Component ablations and spatial uncertainty controls support combining dynamics with aligned uncertainty when allocating a weather transformer's token budget. Code and project page: \href{https://ornl-utome.github.io/}{\textcolor{projectgreen}{\textbf{https://ornl-utome.github.io/}}}.
\end{abstract}

\section{Introduction}
\label{sec:intro}

Data-driven weather models learn forecast operators from reanalysis archives and provide efficient global forecasts~\citep{pathak2022fourcastnet,bi2023accurate,lam2023learning,nguyen2023climax}. Training remains costly for transformer forecasters, which repeatedly process spatial tokens through attention and MLP layers. Uniformly increasing patch size reduces this cost but coarsens every location equally. Adaptive token reduction offers a more selective alternative: it can compress redundant regions while retaining representations of structures relevant to the forecast.

Token pruning and Token Merging (ToMe) reduce sequence length using attention, feature similarity, or learned importance~\citep{rao2021dynamicvit,bolya2022token,bolya2023token}. Weather forecasting introduces an additional consideration: reanalysis inputs estimate a latent atmospheric state using heterogeneous observations and data assimilation~\citep{hersbach2020era5,Janjic2018}. Their uncertainty is spatially structured. Similar hidden features can therefore reflect either reliable atmospheric redundancy or uncertainty in the input estimate, which a similarity-only rule does not distinguish.

We study input reliability as a compute-allocation signal. A latent-input predictive model makes explicit an observation-induced variance component alongside aleatoric variability and epistemic model uncertainty~\citep{KendallGal2017,HullermeierWaegeman2021,McHutchonRasmussen2011}. This motivates considering input uncertainty when allocating tokens, while leaving the allocation rule to empirical evaluation. High uncertainty can coincide with important weather dynamics; uncertainty must therefore be considered jointly with a saliency signal.

\begin{figure}[!t]
    \centering
    \includegraphics[width=0.92\linewidth]{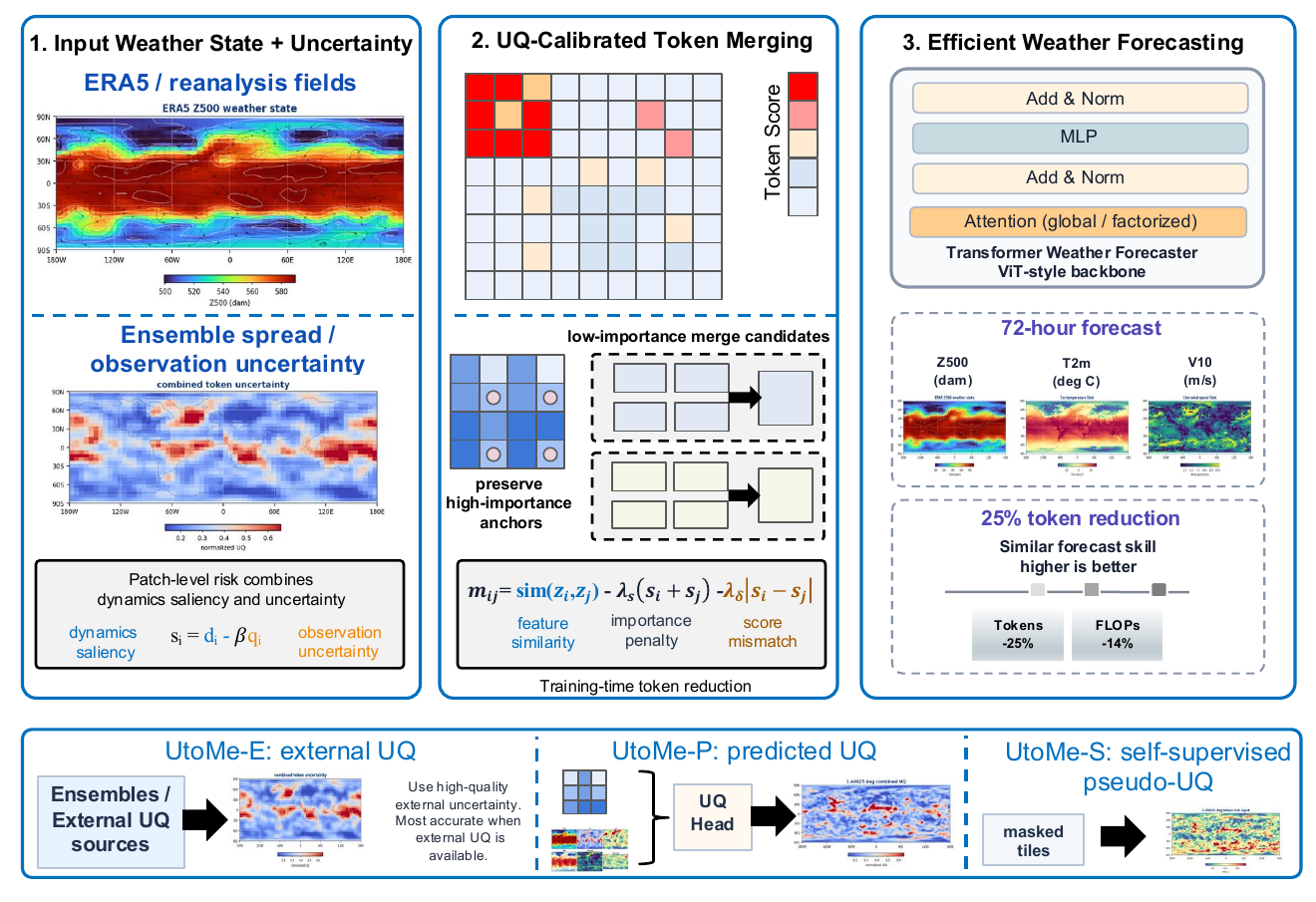}

    \caption{\textbf{\method\ overview.} \method\ uses observation uncertainty to guide training-time token merging for transformer-based weather forecasting. It first turns weather-state saliency and low observation uncertainty into a patch-level token score, then uses that score to protect forecast-sensitive anchors and merge lower-importance redundant tokens. The framework supports three uncertainty-availability regimes: external uncertainty at both stages, during training only, or neither stage.}
    \label{fig:framework}

\end{figure}

\method\ combines patch-level dynamics saliency with normalized ensemble spread to form an importance score (Fig.~\ref{fig:framework}). It protects high-scoring anchors and augments ToMe-style bipartite matching with importance penalties. Merging is active during both training and evaluation. The contribution is a weather-specific allocation policy that retains ToMe's merge operator and accounts for heterogeneous input reliability.

The three variants address distinct uncertainty-availability regimes. \method-E uses external uncertainty at both stages; \method-P learns an externally supervised importance predictor for deployment without uncertainty fields; and \method-S learns from a local spatial-inconsistency proxy without external uncertainty. We make three contributions:
\begin{enumerate}[leftmargin=*, itemsep=2pt, topsep=1pt]
    \item We formulate observation-uncertainty-guided token allocation for weather transformers, using ensemble spread as a relative input-reliability proxy.
    \item We combine dynamics--uncertainty scoring, anchor protection, and importance-calibrated matching, and isolate their contributions with controlled ablations.
    \item We evaluate three uncertainty-availability regimes on ERA5 72-hour forecasting at $5.625^\circ$ and $1.40625^\circ$, including spatial uncertainty controls, difficult-regime diagnostics, pruning comparisons, and a second transformer backbone.
\end{enumerate}

\section{Related Work}
\label{sec:related}

\textbf{Weather foundation models.} Data-driven weather models learn forecast operators from gridded atmospheric archives such as ERA5~\citep{hersbach2020era5,munoz2021era5}. FourCastNet uses adaptive Fourier neural operators for high-resolution global forecasting~\citep{pathak2022fourcastnet}; Pangu-Weather uses a 3D Earth-specific architecture for medium-range forecasting~\citep{bi2023accurate}; GraphCast uses graph neural networks on the sphere~\citep{lam2023learning}; and ClimaX studies a transformer foundation model that can be pretrained across heterogeneous weather and climate datasets~\citep{nguyen2023climax}. Recent efforts continue to scale these models and related Earth-system foundation models~\citep{wang2025orbit,wang2026global}. These works improve forecast skill and model generality at a fixed tokenization. \method\ is complementary: it asks how to reduce the token sequence during training while preserving the forecast information that matters.

\textbf{Token pruning and merging.} Computational efficiency is a central concern in transformer-based learning and adaptation~\citep{tay2022efficient,xi2026prompt,li2026rethinking,zhang2025dpcore,zhang2026adapting,zhang2026prime,zhang2025ot,jin2026deepjepa,yu2026elsa3d,liu2026wan,zhang2026can,wang2026medical,ouyang2026predicting}. Token pruning, routing, pooling, and merging address this concern by reducing sequence length. DynamicViT learns to remove low-importance tokens progressively~\citep{rao2021dynamicvit}. ToMe merges similar tokens with a lightweight bipartite matching algorithm and can be applied to pretrained ViTs, with extensions to diffusion models and other modalities~\citep{bolya2022token,bolya2023token}. Recent work also explores token routing and sparse computation for diffusion models~\citep{zheng2023fast,krause2025tread,sehwag2025stretching,park2025sprint}. Co-Me uses a confidence predictor to guide token merging for visual geometric transformers~\citep{chen2025comeconfidenceguidedtokenmerging}. Sparse-Reslim brings sparse token routing to weather forecasting: it processes a sampled token subset in intermediate transformer blocks and scatters residual updates back to the full sequence, leaving unselected representations intact~\citep{wang2026less}. \method\ uses dynamics saliency and observation uncertainty from the data source to guide anchor protection and pairwise merging during weather-model training.

\textbf{Uncertainty quantification and observation error.} Deep uncertainty quantification is often organized around aleatoric and epistemic uncertainty~\citep{KendallGal2017,HullermeierWaegeman2021,wu2026quarterly}, with practical estimators based on MC dropout, deep ensembles, evidential models, and deterministic distance-aware networks~\citep{GalGhahramani2016,Lakshminarayanan2017,Sensoy2018,LiuEtAl2020SNGP,zhu2026stress,zhu2026residual}. This two-way split usually assumes the input is observed without error. Errors-in-variables models, input-noise propagation, and unified uncertainty formulations make this assumption explicit and add an input or observation uncertainty term~\citep{Fuller1987,Carroll2006,McHutchonRasmussen2011,GastRoth2018,MartinElster2022,ValdenegroToro2024,xu2026rivatfuse}. In geoscience, observation error and representativeness error are core parts of data assimilation~\citep{Janjic2018,hersbach2020era5}. \method\ brings these ideas into the architecture: observation uncertainty is used not only as a diagnostic quantity, but as a token-level signal for allocating computation.

\section{Method}
\label{sec:method}

\subsection{Forecasting Setup}
\label{sec:setup_method}

Let $x_t \in \R^{C \times H \times W}$ denote a multivariate atmospheric state at time $t$, where $C$ includes pressure-level, surface, and static variables. Given lead time $\Delta$, the model predicts $y_{t+\Delta}$. For a ViT-style weather transformer, the grid is split into non-overlapping patches $\{\mathcal{P}_i\}_{i=1}^{N}$ and embedded into tokens $z^0 \in \R^{N \times d}$. A standard transformer processes all $N$ tokens in every layer. Token merging reduces the sequence by combining selected source tokens into destination tokens. The merge policy only assumes this patch-token interface, so it is not tied to a particular weather-transformer implementation.

The main design question is the merge policy. A similarity-only method merges tokens whose hidden features are close. This is attractive because it is cheap and batchable, but it ignores whether the apparent similarity comes from reliable atmospheric redundancy or from noisy observations. \method\ augments the merge policy with an uncertainty-guided importance score computed before merging.

\subsection{Observation Uncertainty as Analytical Motivation}
\label{sec:analytical}

We model the input as a noisy observation of a latent true atmospheric state,
\begin{equation}
\label{eq:obs-model}
\scalebox{0.92}{$
\displaystyle
    \tilde{\bm{x}} = \bm{x}^* + \bm{\varepsilon}, \qquad
    \bm{\varepsilon} \sim p_{\varepsilon}(\cdot \mid \bm{x}^*)
$}
\end{equation}
For training data $\mathcal{D}$, we assume the latent-input and parameter posteriors factorize as $p(\bm{x}^*,\bm{\theta}\mid\tilde{\bm{x}},\mathcal{D})=p(\bm{x}^*\mid\tilde{\bm{x}})p(\bm{\theta}\mid\mathcal{D})$. With finite conditional second moments, the predictive distribution is
\begin{equation}
\label{eq:predictive}
    p(y \mid \tilde{\bm{x}}, \mathcal{D})
    = \int \int p(y \mid \bm{x}^*, \bm{\theta})\,
    p(\bm{x}^* \mid \tilde{\bm{x}})\,
    p(\bm{\theta} \mid \mathcal{D})\, d\bm{x}^*\, d\bm{\theta}.
\end{equation}
The latent-input formulation motivates accounting for observation reliability when allocating a token budget. If one sets $\bm{x}^*=\tilde{\bm{x}}$, Eq.~\eqref{eq:predictive} reduces to the standard setting and predictive variance splits into aleatoric and epistemic terms~\citep{KendallGal2017,Depeweg2018}. Keeping the latent state explicit and applying the law of total variance twice gives
\begin{equation}
\label{eq:three-way}
    \Var[y \mid \tilde{\bm{x}},\mathcal{D}]
    =
    \E_{\bm{x}^*}\!\E_{\bm{\theta}}\!\left[\Var(y \mid \bm{x}^*, \bm{\theta})\right]
    +
    \E_{\bm{x}^*}\!\Var_{\bm{\theta}}\!\left(\E[y \mid \bm{x}^*, \bm{\theta}]\right)
    +
    \Var_{\bm{x}^*}\!\left(\E[y \mid \bm{x}^*]\right).
\end{equation}
The first two terms describe aleatoric variability and epistemic model uncertainty, averaged over possible latent states. The third term is observation uncertainty: it measures how much the forecast mean changes because several latent states could have produced the same observed input. Appendix~\ref{app:proof} gives the derivation and the first-order approximation that connects this term to local input sensitivity and latent-input posterior covariance.

The decomposition motivates including input reliability in token allocation; it does not imply an optimal compression policy. We use ensemble spread as a relative proxy for uncertainty in the input estimate and combine it with dynamics saliency. The subtraction sign, anchor rule, and matching penalties are evaluated through ablations rather than derived from the variance identity.

\subsection{Patch-Level Importance from Dynamics and Observation Spread}
\label{sec:risk}

For variables with uncertainty information, we use an aligned field $u_t\in\R^{C\times H\times W}$ from ECMWF's Ensemble of Data Assimilations (EDA). Its spread is a relative estimate of reanalysis uncertainty, including observation and assimilation-model effects; it is not an isolated measurement of observation-error variance.\footnote{ECMWF/Copernicus ERA5 uncertainty estimation documentation: \url{https://confluence.ecmwf.int/display/CKB/ERA5:+uncertainty+estimation}.} We refer to this relative input-reliability signal as observation uncertainty and abbreviate uncertainty quantification as UQ. Low spread strengthens protection when a patch is dynamically salient; a patch with high spread can also remain important when its dynamics signal is large. In our reported runs, the UQ tensor follows the input-channel layout, with static variables assigned zero uncertainty. We normalize each variable and patch-average the field:
\begin{equation}
\label{eq:qpatch}
    q_i =
    \frac{1}{|\mathcal{P}_i|\,C}
    \sum_{c=1}^{C}
    \sum_{(h,w)\in\mathcal{P}_i}
    \hat{u}_{t,c,h,w},
\end{equation}
where $\hat{u}$ denotes the normalized spread and $\mathcal{P}_i$ is the spatial support of token $i$. We then standardize the patch scores per sample and keep the notation $q_i$ for the normalized patch-level observation-uncertainty score.

Dynamics saliency combines spatial variation and anomaly amplitude in the normalized input-time fields of the five forecast variables. For their channel set $\mathcal{C}_{\mathrm{dyn}}$, define
\begin{equation}
\label{eq:dynamics}
 D_{h,w}=\frac{1}{|\mathcal{C}_{\mathrm{dyn}}|}\sum_{c\in\mathcal{C}_{\mathrm{dyn}}}
 \left(|\delta_h x_{t,c,h,w}|+|\delta_w x_{t,c,h,w}|+
 \tfrac12|x_{t,c,h,w}-\bar{x}_{t,c}|\right),
\end{equation}
where $\delta_h$ and $\delta_w$ are adjacent-grid differences and $\bar{x}_{t,c}$ is the spatial mean of input channel $c$. Patch averaging followed by per-sample standardization gives $d_i$. This construction emphasizes sharp gradients and anomalies using information available at the input time. The final importance score is
\begin{equation}
\label{eq:risk}
    s_i = d_i - \beta q_i,
\end{equation}
with weight $\beta$. We use $\beta=1$, selected on validation data: dynamics saliency increases merge protection, while observation spread subtracts from the score. Thus, high-dynamics low-spread patches receive the strongest protection, and high-spread patches become better merge candidates unless the dynamics signal is strong enough to dominate. Fig.~\ref{fig:highres_case_study} illustrates the spatial structure retained by patch aggregation.

\subsection{Anchored, Uncertainty-Calibrated Token Merging}
\label{sec:merging}

\method\ keeps the efficiency and batchability of ToMe-style bipartite matching, but changes the merge decision so that it respects the importance score in Eq.~\eqref{eq:risk}. Let $\mathcal{A}$ contain the top $\rho$ fraction of tokens according to $s_i$, with $\rho=0.25$ by default. Tokens in $\mathcal{A}$ are treated as anchors: they retain separate representations and are excluded from both merge-source and merge-destination pools. The anchor fraction is applied when merging begins, and the protection mask is carried through subsequent layers.

After anchors are removed from both candidate pools, \method\ applies an uncertainty-calibrated matching score to the remaining tokens:
\begin{equation}
\label{eq:match}
    m_{ij}
    =
    \mathrm{sim}(z_i, z_j)
    -
    \lambda_s(s_i+s_j)
    -
    \lambda_{\Delta}|s_i-s_j|,
\end{equation}
We use $\lambda_s=\lambda_{\Delta}=0.05$, with the selection protocol and sensitivity analysis in Appendix~\ref{app:ablations}. Here $z_i$ and $z_j$ are hidden tokens, $\mathrm{sim}$ is cosine similarity in our implementation, and $\lambda_s,\lambda_{\Delta}\ge 0$ control the strength of the importance penalties. The similarity term keeps the original ToMe criterion: if two tokens encode the same local atmospheric information, they remain good merge candidates. The sum penalty discourages high-importance pairs, while the difference penalty discourages pairs with dissimilar joint importance. For each source token $i$, the destination is selected as
\begin{equation}
\label{eq:destination}
    j^*(i) = \argmax_{j\in\mathcal{D}_{\mathrm{dst}}} m_{ij},
\end{equation}
where $\mathcal{D}_{\mathrm{dst}}$ is the destination set in the bipartite partition. The top-scoring edges are merged until the layer budget is reached, and merged token features are averaged with size tracking as in ToMe.

\subsection{Variants for Different UQ Availability Regimes}
\label{sec:variants}

\textbf{\method-E: external UQ at training and inference.} \method-E uses the EDA-derived uncertainty field in Eq.~\eqref{eq:qpatch} at both stages. This is the cleanest setting for reanalysis workflows where ensemble spread is available alongside the input fields.

\textbf{\method-P: external UQ during training only.} \method-P handles the case where uncertainty fields are available during training but not at deployment. We add a lightweight predictor $g_{\phi}$ that maps early hidden tokens to scores $\hat{s}_i$. During training, $g_{\phi}$ is supervised by the EDA-derived score $s_i$ with a pairwise-ranking loss, since the merge policy depends primarily on the ordering used for anchor selection and edge penalties. Ranking supervision targets anchor selection; the matching penalties additionally depend on score magnitudes. At inference, $\hat{s}_i$ replaces $s_i$ in anchor selection and matching.

\textbf{\method-S: no external UQ.} \method-S removes the dependency on ensemble products. We build a pseudo-uncertainty target from local spatial inconsistency,
\begin{equation}
\label{eq:pseudo}
\scalebox{0.92}{$
\displaystyle
    \tilde{q}_{h,w}
    =
    \frac{1}{|\mathcal{C}_{\mathrm{dyn}}|}\sum_{c\in\mathcal{C}_{\mathrm{dyn}}}
    \left|x_{t,c,h,w} - \mathrm{AvgPool}_{3\times 3}(x_{t,c})_{h,w}\right|
$}
\end{equation}
using the same input-time variable set as dynamics saliency. We patch-average and standardize this proxy, substitute it for $q_i$ in Eq.~\eqref{eq:risk}, and train the same predictor used by \method-P on the resulting importance score. The proxy measures local inconsistency and is not a calibrated observation-error estimate.

\FloatBarrier
\section{Experiments}
\label{sec:experiments}
\suppressfloats[t]

\subsection{Experimental Setup}
\label{sec:setup}

\textbf{Data and task.} We evaluate global ERA5 deterministic forecasting at two resolutions. The $5.625^\circ$ setting uses 2010 to 2014 for training, 2015 for validation, and 2016 for testing, with a $32\times64$ grid and $2\times2$ patches, giving $512$ initial spatial tokens. The $1.40625^\circ$ setting uses 1979 to 2018 for training, 2019 for validation, and 2020 for testing, with a $128\times256$ grid and $4\times4$ patches, giving $2048$ initial spatial tokens. These angles specify the latitude--longitude grid spacing. Each sample uses the atmospheric state $x_t$ to predict selected targets at $t+72$ hours. The input features use the ERA5 variable set from ClimaX and contain $41$ channels covering static fields, surface variables, and pressure-level variables; the output features are Z500 (500 hPa geopotential, $\mathrm{m^2\,s^{-2}}$), T850 (850 hPa temperature, K), T2m (2 m temperature, K), and U10/V10 (zonal/meridional 10 m winds, $\mathrm{m\,s^{-1}}$). These targets cover mid-tropospheric circulation, lower-tropospheric temperature, near-surface temperature, and surface winds. EDA spread is aligned with the input time, grid, and variables where available. It is not concatenated to the forecast input; it supervises or calibrates token merging.

\textbf{Model and baselines.} The main comparisons use the same ViT-style transformer forecaster: eight transformer layers, hidden dimension $1024$, $16$ attention heads, MLP ratio $4$, dropout $0.1$, drop-path $0.1$, and a two-layer decoder. The model embeds each selected variable with a one-channel patch embed, adds variable embeddings, aggregates variables into a spatial token sequence, and applies token merging on that spatial sequence. We compare no merging, train-time ToMe with the same layerwise token budget, and \method-E/P/S. For all merging rows, merging is active during both training and evaluation. E and ToMe share a schedule; P/S delay merging until after the predictor probe (Appendix~\ref{app:impl}). The main tables report final token reductions of $25\%$ and $50\%$.

\textbf{Training and metrics.} Training uses DDP on an HPC cluster of 8 nodes, each exposing $8$ AMD MI250X GPU devices, for $64$ GPU devices in total. We optimize with AdamW using a peak learning rate of $5\times10^{-4}$, weight decay $10^{-5}$, $1000$ warmup steps, gradient clipping at $1.0$, and a $12000$-step learning-rate schedule. The per-process batch size is $16$ at $5.625^\circ$ and $1$ at $1.40625^\circ$. Predictor-based variants use an auxiliary pairwise-ranking loss with weight $0.2$. We report mean and standard deviation over three random seeds for the main forecast comparisons; the additional controls report point estimates. Metrics are latitude-weighted root mean square error (wRMSE) for each target variable, Overall error aggregated over the targets, and anomaly correlation coefficient (ACC) after climatology removal, following ClimaX~\citep{nguyen2023climax}. Compute is the encoder FLOPs ratio relative to no merging at the same resolution. We use normalized FLOPs as the main efficiency metric because it is controlled by the token schedule and is comparable across resolutions. More details are in Appendix~\ref{app:impl}.

Overall is the arithmetic mean of the five target-specific errors:
\begin{equation}
\label{eq:overall}
\mathrm{Overall}=\frac{1}{5}\sum_{k\in\{\mathrm{Z500,T850,T2m,U10,V10}\}}\mathrm{wRMSE}_k.
\end{equation}
Overall mixes physical units and is numerically dominated by Z500; it is not a scale-invariant measure of multivariate skill. We therefore interpret it alongside per-variable wRMSE and ACC.

\begin{figure}[!t]
    \centering
    \includegraphics[width=0.90\linewidth]{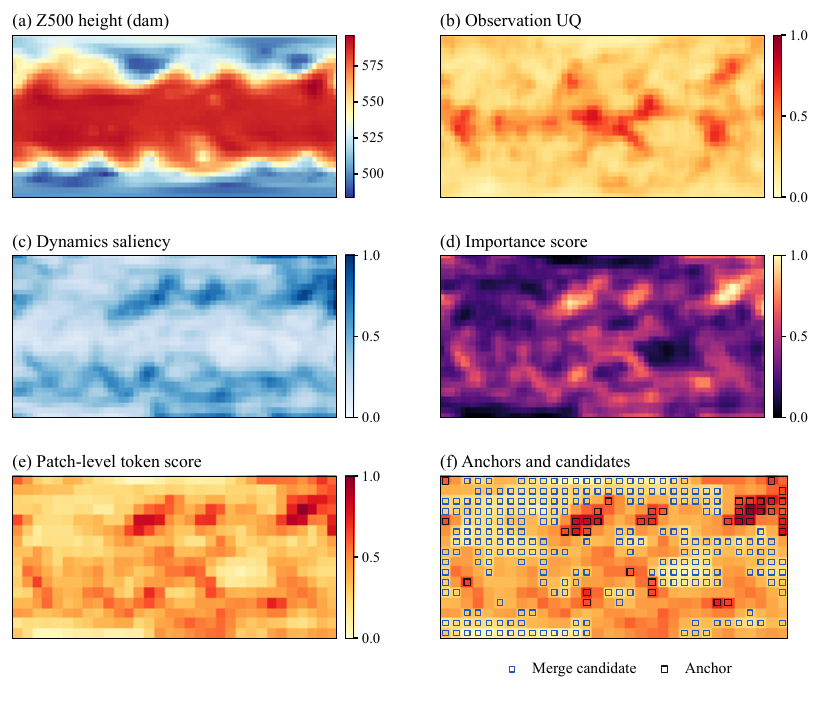}

    \caption{\textbf{$5.625^\circ$ case summary.} The figure shows the complete merge-decision pipeline: weather state, observation uncertainty, dynamics saliency, importance score, patch-level token score, and the final anchor or merge-candidate overlay.}
    \label{fig:summary_panel_5p625}

\end{figure}

\begin{table}[htbp]
    \centering
    \scriptsize
    \setlength{\tabcolsep}{1.6pt}
    \caption{Main results on \textbf{$5.625^\circ$} ERA5 72-hour forecasting. The table compares no merging, train-time ToMe, and the three \method\ variants under matched final token-reduction ratios. Values are mean and standard deviation over three random seeds. Lower is better for Overall and per-variable wRMSE, higher is better for ACC, and Compute is the encoder FLOPs ratio relative to no merging at the same resolution; the main resolution tables follow this convention. Green shades identify E/P/S uncertainty availability; blue bands identify token budgets. Bold/underlined values are best/second-best within each budget.}
    \label{tab:main_5p625}
    \resizebox{\linewidth}{!}{%
    \begin{tabular}{@{}lcccccccr@{}}
        \toprule
        Method & Overall $\downarrow$ & ACC $\uparrow$ & Z500 $\downarrow$ & T850 $\downarrow$ & T2m $\downarrow$ & U10 $\downarrow$ & V10 $\downarrow$ & Compute $\downarrow$ \\
        \midrule
        \refval{No merging} & \refval{$143.07_{\pm 1.81}$} & \refval{$0.651_{\pm 0.008}$} & \refval{$701.64_{\pm 8.92}$} & \refval{$3.108_{\pm 0.023}$} & \refval{$2.502_{\pm 0.008}$} & \refval{$3.944_{\pm 0.024}$} & \refval{$4.151_{\pm 0.051}$} & \refval{$1.00\times$} \\
        \midrule
        \rowcolor{tokenblue25}
        \multicolumn{9}{l}{\textit{25\% token reduction}} \\
        ToMe & $160.82_{\pm 2.07}$ & $0.562_{\pm 0.011}$ & $789.00_{\pm 10.34}$ & $3.544_{\pm 0.041}$ & $3.378_{\pm 0.038}$ & $\mathbf{4.041}_{\pm 0.047}$ & $\mathbf{4.142}_{\pm 0.044}$ & $0.86\times$ \\
        \rowcolor{utomeEbg}
        \method-E & $\underline{144.74}_{\pm 1.42}$ & $\underline{0.640}_{\pm 0.007}$ & $\underline{709.50}_{\pm 7.21}$ & $\underline{3.220}_{\pm 0.026}$ & $\underline{2.730}_{\pm 0.021}$ & $4.060_{\pm 0.032}$ & $4.190_{\pm 0.039}$ & $0.86\times$ \\
        \rowcolor{utomePbg}
        \method-P & $\mathbf{144.10}_{\pm 1.16}$ & $\mathbf{0.650}_{\pm 0.006}$ & $\mathbf{706.34}_{\pm 6.43}$ & $\mathbf{3.210}_{\pm 0.024}$ & $\mathbf{2.717}_{\pm 0.019}$ & $\underline{4.054}_{\pm 0.030}$ & $\underline{4.182}_{\pm 0.034}$ & $0.87\times$ \\
        \rowcolor{utomeSbg}
        \method-S & $145.36_{\pm 1.69}$ & $0.635_{\pm 0.009}$ & $712.50_{\pm 8.18}$ & $3.260_{\pm 0.031}$ & $2.760_{\pm 0.024}$ & $4.075_{\pm 0.036}$ & $4.205_{\pm 0.041}$ & $0.87\times$ \\
        \midrule
        \rowcolor{tokenblue50}
        \multicolumn{9}{l}{\textit{50\% token reduction}} \\
        ToMe & $220.05_{\pm 3.84}$ & $0.423_{\pm 0.014}$ & $1080.19_{\pm 18.76}$ & $5.189_{\pm 0.092}$ & $5.822_{\pm 0.117}$ & $4.586_{\pm 0.066}$ & $4.464_{\pm 0.058}$ & $0.55\times$ \\
        \rowcolor{utomeEbg}
        \method-E & $\mathbf{151.95}_{\pm 1.58}$ & $\mathbf{0.605}_{\pm 0.010}$ & $\mathbf{744.41}_{\pm 7.82}$ & $\underline{3.466}_{\pm 0.034}$ & $\underline{3.485}_{\pm 0.036}$ & $\underline{4.125}_{\pm 0.039}$ & $\underline{4.255}_{\pm 0.042}$ & $0.55\times$ \\
        \rowcolor{utomePbg}
        \method-P & $\underline{152.42}_{\pm 1.47}$ & $\underline{0.580}_{\pm 0.012}$ & $\underline{747.19}_{\pm 7.31}$ & $\mathbf{3.412}_{\pm 0.029}$ & $\mathbf{3.481}_{\pm 0.033}$ & $\mathbf{3.963}_{\pm 0.028}$ & $\mathbf{4.045}_{\pm 0.035}$ & $0.57\times$ \\
        \rowcolor{utomeSbg}
        \method-S & $167.63_{\pm 2.26}$ & $0.502_{\pm 0.015}$ & $821.37_{\pm 11.94}$ & $3.732_{\pm 0.047}$ & $4.507_{\pm 0.071}$ & $4.196_{\pm 0.044}$ & $4.335_{\pm 0.049}$ & $0.57\times$ \\
        \bottomrule
    \end{tabular}}
\end{table}

\subsection{Forecast Skill at Matched Token Budgets}
\label{sec:main_results}

\begin{table}[!t]
    \centering
    \scriptsize
    \setlength{\tabcolsep}{1.6pt}
    \caption{Main results on \textbf{$1.40625^\circ$} ERA5 72-hour forecasting. Values are mean and standard deviation over three seeds; conventions follow Table~\ref{tab:main_5p625}.}
    \label{tab:main_1p4}
    \resizebox{\linewidth}{!}{%
    \begin{tabular}{@{}lcccccccr@{}}
        \toprule
        Method & Overall $\downarrow$ & ACC $\uparrow$ & Z500 $\downarrow$ & T850 $\downarrow$ & T2m $\downarrow$ & U10 $\downarrow$ & V10 $\downarrow$ & Compute $\downarrow$ \\
        \midrule
        \refval{No merging} & \refval{$32.41_{\pm 0.28}$} & \refval{$0.964_{\pm 0.004}$} & \refval{$156.76_{\pm 1.34}$} & \refval{$1.299_{\pm 0.007}$} & \refval{$1.269_{\pm 0.006}$} & \refval{$1.339_{\pm 0.008}$} & \refval{$1.360_{\pm 0.009}$} & \refval{$1.00\times$} \\
        \midrule
        \rowcolor{tokenblue25}
        \multicolumn{9}{l}{\textit{25\% token reduction}} \\
        ToMe & $42.45_{\pm 0.24}$ & $0.949_{\pm 0.003}$ & $206.02_{\pm 1.17}$ & $1.452_{\pm 0.006}$ & $1.483_{\pm 0.007}$ & $1.572_{\pm 0.008}$ & $1.724_{\pm 0.009}$ & $0.86\times$ \\
        \rowcolor{utomeEbg}
        \method-E & $\mathbf{39.59}_{\pm 0.21}$ & $\mathbf{0.952}_{\pm 0.003}$ & $\mathbf{191.93}_{\pm 1.05}$ & $\mathbf{1.413}_{\pm 0.005}$ & $\mathbf{1.454}_{\pm 0.006}$ & $\mathbf{1.527}_{\pm 0.007}$ & $\mathbf{1.645}_{\pm 0.008}$ & $0.86\times$ \\
        \rowcolor{utomePbg}
        \method-P & $\underline{39.61}_{\pm 0.01}$ & $\mathbf{0.952}_{\pm 0.000}$ & $\underline{191.98}_{\pm 0.03}$ & $1.425_{\pm 0.001}$ & $1.459_{\pm 0.000}$ & $\underline{1.530}_{\pm 0.000}$ & $\underline{1.648}_{\pm 0.001}$ & $0.87\times$ \\
        \rowcolor{utomeSbg}
        \method-S & $39.84_{\pm 0.27}$ & $\underline{0.951}_{\pm 0.004}$ & $193.14_{\pm 1.31}$ & $\underline{1.416}_{\pm 0.007}$ & $\underline{1.458}_{\pm 0.008}$ & $1.533_{\pm 0.009}$ & $1.656_{\pm 0.010}$ & $0.87\times$ \\
        \midrule
        \rowcolor{tokenblue50}
        \multicolumn{9}{l}{\textit{50\% token reduction}} \\
        ToMe & $67.32_{\pm 0.43}$ & $0.905_{\pm 0.005}$ & $328.21_{\pm 2.11}$ & $1.846_{\pm 0.011}$ & $2.056_{\pm 0.015}$ & $2.136_{\pm 0.014}$ & $\underline{2.347}_{\pm 0.018}$ & $0.55\times$ \\
        \rowcolor{utomeEbg}
        \method-E & $\mathbf{53.00}_{\pm 0.36}$ & $\mathbf{0.913}_{\pm 0.004}$ & $\mathbf{256.72}_{\pm 1.76}$ & $\underline{1.759}_{\pm 0.009}$ & $\underline{2.051}_{\pm 0.012}$ & $\underline{2.038}_{\pm 0.013}$ & $2.433_{\pm 0.015}$ & $0.55\times$ \\
        \rowcolor{utomePbg}
        \method-P & $\underline{53.14}_{\pm 0.01}$ & $0.907_{\pm 0.000}$ & $\underline{257.37}_{\pm 0.05}$ & $1.762_{\pm 0.001}$ & $2.056_{\pm 0.001}$ & $2.045_{\pm 0.001}$ & $2.444_{\pm 0.001}$ & $0.57\times$ \\
        \rowcolor{utomeSbg}
        \method-S & $57.01_{\pm 0.04}$ & $\underline{0.908}_{\pm 0.000}$ & $276.99_{\pm 0.18}$ & $\mathbf{1.752}_{\pm 0.001}$ & $\mathbf{1.968}_{\pm 0.004}$ & $\mathbf{2.009}_{\pm 0.002}$ & $\mathbf{2.335}_{\pm 0.002}$ & $0.57\times$ \\
        \bottomrule
    \end{tabular}}
\end{table}

\textbf{Joint scoring improves the accuracy--compute trade-off.} Tables~\ref{tab:main_5p625} and~\ref{tab:main_1p4} compare allocation policies at matched final token budgets. At $5.625^\circ$ and $25\%$ reduction, \method-P obtains Overall error $144.10$, compared with $143.07$ without merging and $160.82$ for ToMe. At $50\%$, \method-E obtains $151.95$, compared with ToMe's $220.05$. At $1.40625^\circ$, \method-E lowers ToMe's Overall error by $2.86$ and $14.32$ at the two budgets, respectively, at identical encoder FLOPs. These gains do not establish parity with the full-token model: at high resolution and $25\%$ reduction, Z500 wRMSE increases from $156.76$ to $191.93$ ($22.4\%$). Nor are the gains uniform across targets: ToMe has lower U10/V10 errors at low-resolution $25\%$ reduction and lower V10 error than E/P at high-resolution $50\%$ reduction.

\takeawaypoint{Takeaway 1: Joint scoring improves budgeted forecasting}{At matched token budgets, \method\ lowers Overall error relative to ToMe at both resolutions. The per-variable results identify the benefits and costs of reallocating the token budget.}

\begin{figure}[!t]
    \centering
    \begin{subfigure}[t]{0.455\linewidth}
        \centering
        \includegraphics[width=\linewidth,height=0.20\textheight,keepaspectratio]{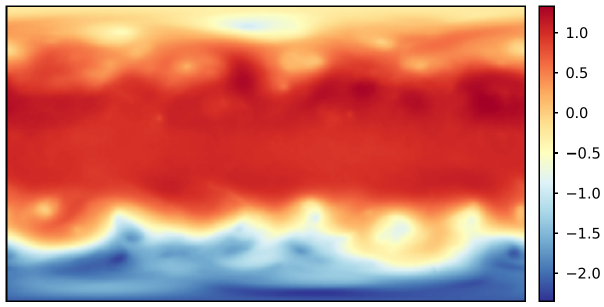}
        \caption{\textbf{Z500 input.}}
    \end{subfigure}
    \hfill
    \begin{subfigure}[t]{0.455\linewidth}
        \centering
        \includegraphics[width=\linewidth,height=0.20\textheight,keepaspectratio]{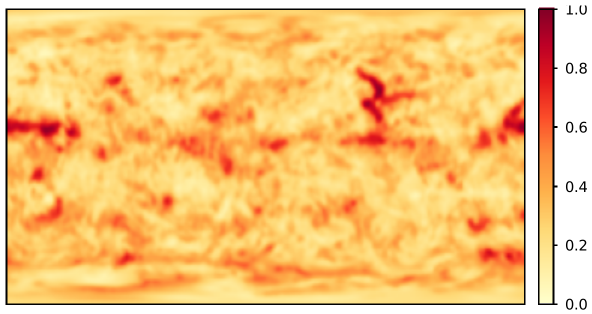}
        \caption{\textbf{Observation uncertainty.}}
    \end{subfigure}

    \begin{subfigure}[t]{0.455\linewidth}
        \centering
        \includegraphics[width=\linewidth,height=0.20\textheight,keepaspectratio]{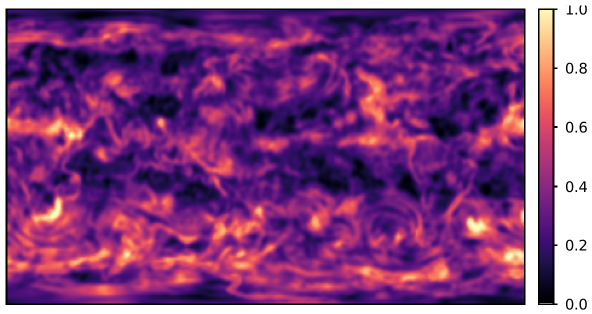}
        \caption{\textbf{Importance score.}}
    \end{subfigure}
    \hfill
    \begin{subfigure}[t]{0.455\linewidth}
        \centering
        \includegraphics[width=\linewidth,height=0.20\textheight,keepaspectratio]{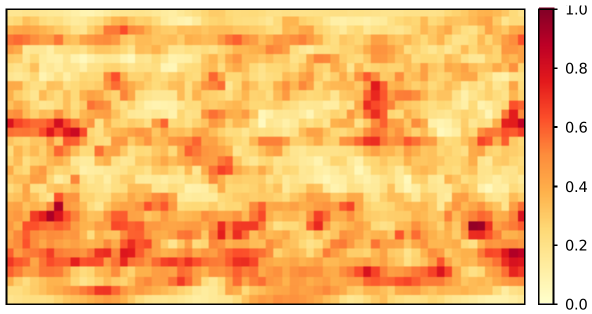}
        \caption{\textbf{Patch-level token score.}}
    \end{subfigure}

    \caption{\textbf{$1.40625^\circ$ high-resolution case study.} The observation-uncertainty field contains coherent bands and high-spread structures. After saliency combination and patch aggregation, these structures produce a heterogeneous token score for uncertainty-guided merging.}
    \label{fig:highres_case_study}
\end{figure}

\textbf{Token-reduction ratio ablation.}
\label{sec:ratio}
Table~\ref{tab:ablation_ratio} quantifies the high-resolution accuracy--compute frontier. At $5\%$ reduction, Overall error rises from $32.41$ to $32.80$ ($1.2\%$), ACC remains $0.964$, and encoder FLOPs decrease by $3\%$. At $10\%$, Overall error rises by $6.4\%$ for a $7\%$ FLOPs reduction. At $25\%$, the $14\%$ FLOPs saving comes with a $22.2\%$ increase in Overall error. The $50\%$ setting saves $45\%$ of encoder FLOPs but increases Overall error by $63.5\%$. Thus, the low-loss operating point is the small-reduction regime; the larger budgets test robustness to substantial compression.

\takeawaypoint{Takeaway 2: Compression has a measurable accuracy cost}{Small reductions preserve most full-token skill. At $25\%$ and $50\%$ reduction, \method\ improves the trade-off relative to ToMe while incurring material losses relative to no merging.}

\textbf{Complementary uncertainty regimes.} At high resolution and $50\%$ reduction, \method-S has lower errors than E/P on T850, T2m, U10, and V10, while \method-E is stronger on Z500, Overall, and ACC. The mixed result is consistent with complementary allocation signals and does not establish that external spread is preferable for every target. Z500 dominates the numerical scale of Overall.

\textbf{Additional baselines and backbone.} With the joint score held fixed, \method-E merging improves on score-based pruning at both budgets. On a plain 12-layer ViT, it improves over ToMe by $2.7\%$/$4.9\%$ at matched FLOPs. Appendix~\ref{app:baselines} reports the complete comparisons.

\begin{table}[!t]
    \centering
    \scriptsize
    \setlength{\tabcolsep}{1.6pt}
    \caption{Token-reduction ratio ablation at \textbf{$1.40625^\circ$}. \method\ corresponds to \method-E.
    }
    \label{tab:ablation_ratio}
    \resizebox{\linewidth}{!}{%
    \begin{tabular}{@{}llcccccccr@{}}
        \toprule
        Method & Token Red. & Overall $\downarrow$ & ACC $\uparrow$ & Z500 $\downarrow$ & T850 $\downarrow$ & T2m $\downarrow$ & U10 $\downarrow$ & V10 $\downarrow$ & Compute $\downarrow$ \\
        \midrule
        \refval{No merging} & \refval{$0\%$} & \refval{$32.41_{\pm 0.28}$} & \refval{$0.964_{\pm 0.004}$} & \refval{$156.76_{\pm 1.34}$} & \refval{$1.299_{\pm 0.007}$} & \refval{$1.269_{\pm 0.006}$} & \refval{$1.339_{\pm 0.008}$} & \refval{$1.360_{\pm 0.009}$} & \refval{$1.00\times$} \\
        \midrule
        \rowcolor{tokenblue05}
        \method & $5\%$  & $32.80_{\pm 0.18}$ & $0.964_{\pm 0.002}$ & $158.70_{\pm 0.94}$ & $1.305_{\pm 0.005}$ & $1.278_{\pm 0.005}$ & $1.349_{\pm 0.006}$ & $1.372_{\pm 0.007}$ & $0.97\times$ \\
        \rowcolor{tokenblue10}
        \method & $10\%$ & $34.49_{\pm 0.25}$ & $0.961_{\pm 0.003}$ & $167.00_{\pm 1.16}$ & $1.330_{\pm 0.006}$ & $1.310_{\pm 0.006}$ & $1.380_{\pm 0.007}$ & $1.410_{\pm 0.008}$ & $0.93\times$ \\
        \rowcolor{tokenblue25}
        \method & $25\%$ & $39.59_{\pm 0.21}$ & $0.952_{\pm 0.003}$ & $191.93_{\pm 1.05}$ & $1.413_{\pm 0.005}$ & $1.454_{\pm 0.006}$ & $1.527_{\pm 0.007}$ & $1.645_{\pm 0.008}$ & $0.86\times$ \\
        \rowcolor{tokenblue50}
        \method & $50\%$ & $53.00_{\pm 0.36}$ & $0.913_{\pm 0.004}$ & $256.72_{\pm 1.76}$ & $1.759_{\pm 0.009}$ & $2.051_{\pm 0.012}$ & $2.038_{\pm 0.013}$ & $2.433_{\pm 0.015}$ & $0.55\times$ \\
        \bottomrule
    \end{tabular}}
\end{table}
\textbf{Uncertainty predictor ablation.}
\label{sec:predictor}
\method-P and \method-S obtain forecasting gains without external uncertainty at inference (Tables~\ref{tab:main_5p625} and~\ref{tab:main_1p4}). On three held-out $5.625^\circ$ cases, P's predicted scores have Pearson correlations $0.769$--$0.825$ with the EDA-derived teacher, while S reaches $0.858$--$0.900$ against its distinct proxy target (Appendix~\ref{app:predictor_quality}). These correlations describe score agreement, not uncertainty calibration, and are not directly comparable across targets. Ranking supports anchor selection, whereas matching also uses score magnitudes. The transformation controls below test empirical robustness to selected magnitude changes.

\takeawaypoint{Takeaway 3: External spread is optional at deployment}{P distills an externally supervised score; S learns a local-inconsistency proxy. Both support merging without external uncertainty at inference, with target-dependent differences in forecast skill.}

\begin{figure}[!t]
    \centering
    \includegraphics[width=0.90\linewidth,height=0.28\linewidth,keepaspectratio]{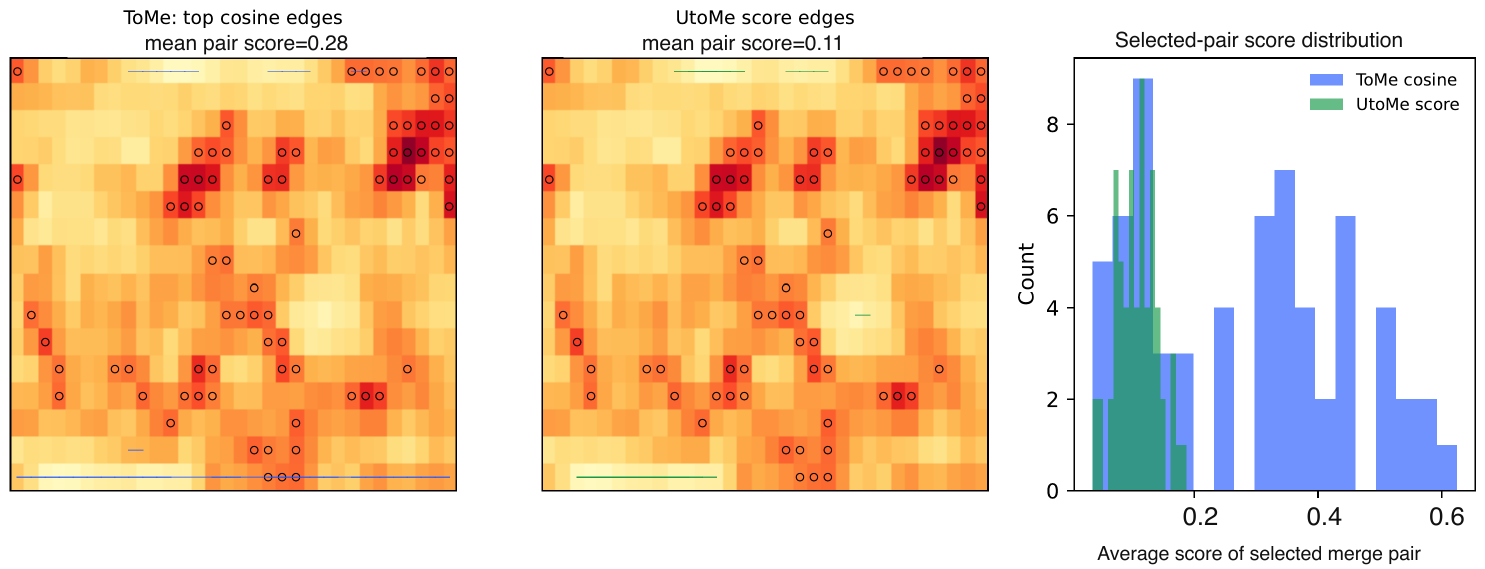}

    \caption{\textbf{Cosine-only matching can select merge edges in high-importance regions because feature similarity alone does not encode observation uncertainty.} \method\ protects anchors and penalizes importance-incompatible pairs, shifting the merge budget toward lower-importance regions.}
    \label{fig:cosine_failure}

\end{figure}

\textbf{Qualitative Analysis}
\label{sec:qualitative}
Fig.~\ref{fig:summary_panel_5p625} shows the low-resolution allocation pipeline; Fig.~\ref{fig:highres_case_study} illustrates its high-resolution signals. Fig.~\ref{fig:cosine_failure} compares selected edges under cosine-only and joint-importance matching. At $25\%$ reduction, the merge-edge Jaccard overlap is $0.139$, and merged-token membership in the top importance quartile falls from $46.8\%$ to $0\%$. These diagnostics establish that the policy changes selected tokens; the forecasting comparisons establish its utility. Appendix~\ref{app:merge_decisions} distinguishes quantitative statistics from illustrative figure markers.

\subsection{Component Contributions and UQ Robustness}
\label{sec:components}

\begin{table}[!htbp]
\centering
\small
\setlength{\tabcolsep}{7pt}
\caption{\textbf{UQ, anchors, and matching contribute at both budgets.} Overall wRMSE ($\downarrow$) at $1.40625^\circ$ with matched token schedules.}
\label{tab:components_main}
\begin{tabular}{@{}lrr@{}}
\toprule
Configuration & 25\% reduction & 50\% reduction \\
\midrule
    ToMe & 42.45 & 67.32 \\
    Dynamics only ($\beta=0$) & 40.95 & 58.60 \\
    No anchors & 40.31 & 61.40 \\
    No UQ-aware matching & 40.68 & 55.90 \\
    \midrule
    \textbf{Full \method-E} & \textbf{39.59} & \textbf{53.00} \\
\bottomrule
\end{tabular}
\end{table}
\begin{table}[!htbp]
\centering
\footnotesize
\setlength{\tabcolsep}{5pt}
\caption{\textbf{Spatial alignment matters more than the tested rescalings.} Score rank correlation and anchor overlap are relative to the EDA-derived score; errors are Overall ($\downarrow$) at $1.40625^\circ$. The two-member row uses simulated sampling noise.}
\label{tab:uq_robust_main}
\begin{tabular}{@{}lrrrr@{}}
\toprule
UQ field & Rank corr. & Anchor overlap & 25\% & 50\% \\
\midrule
Original EDA & 1.000 & 100\% & 39.59 & 53.00 \\
$\sqrt{\mathrm{UQ}}$ & 0.995 & 94\% & 39.62 & 53.05 \\
$\mathrm{UQ}^2$ & 0.977 & 88\% & 39.68 & 53.19 \\
Rank-mapped UQ & 0.969 & 84\% & 39.72 & 53.28 \\
Simulated 2-member & 0.960 & 79\% & 39.94 & 54.10 \\
Constant UQ & -- & -- & 40.95 & 58.60 \\
Spatially shuffled UQ & $\approx0$ & 26\% & 41.88 & 62.30 \\
\bottomrule
\end{tabular}
\end{table}

\textbf{Contribution of the score, anchors, and matching.} In Table~\ref{tab:components_main}, removing UQ increases Overall wRMSE by 1.36/5.60 at $25\%$/$50\%$ reduction. Removing anchors costs 0.72/8.40, and removing the matching penalties costs 1.09/2.90. The full component matrix in Appendix~\ref{app:ablations} also shows that UQ-only scoring underperforms the combined score.

\textbf{Spatial information and rescaling.} In Table~\ref{tab:uq_robust_main}, shuffling UQ degrades wRMSE to 41.88/62.30. Square-root, square, and rank transformations change wRMSE by at most 0.13/0.28, indicating that preserving spatial information matters more than these magnitude transformations. Appendix~\ref{app:uq_robustness} reports the full controls and the finite-ensemble sampling-noise simulation.

\subsection{Forecast Quality in Difficult Regimes}
\label{sec:difficult_regimes}

\begin{table}[!htbp]
\centering
\footnotesize
\setlength{\tabcolsep}{5pt}
\caption{\textbf{The gain concentrates in dynamically active regions, including high-UQ regions.} Z500 wRMSE ($\downarrow$) at $1.40625^\circ$ and $25\%$ token reduction. L/H denote below/at-or-above the per-sample median of UQ ($q$) and dynamics ($d$). The two low-dynamics regimes expose the allocation trade-off.}
\label{tab:regimes_main}
\begin{tabular}{@{}lrrrrr@{}}
\toprule
Method & $q$L/$d$H & $q$H/$d$H & $q$L/$d$L & $q$H/$d$L & Global \\
\midrule
    No merging & 172.0 & 200.0 & 116.0 & 124.0 & 156.76 \\
    ToMe & 219.8 & 268.4 & 158.2 & 155.7 & 206.02 \\
    \method-E & 185.2 & 220.0 & 176.0 & 181.0 & 191.93 \\
\bottomrule
\end{tabular}
\end{table}

\textbf{High uncertainty with strong dynamics.} Using per-sample median splits of UQ and dynamics, Table~\ref{tab:regimes_main} shows that \method-E lowers high-UQ/high-dynamics Z500 wRMSE from 268.4 to 220.0, an $18.0\%$ improvement over ToMe at $25\%$ reduction. It also improves the low-UQ/high-dynamics regime, while errors increase in the two low-dynamics regimes. The result supports joint allocation in this regime, while exposing a cost in less dynamically active regions.

\textbf{Extremes and spatial structure.} At $50\%$ reduction, \method-E lowers upper-tail Z500, lower-tail Z500, and upper-tail T2m errors by $22.4\%$, $22.7\%$, and $16.9\%$ relative to ToMe. Spatial standard-deviation, gradient, kinetic-energy, and vorticity ratios also remain closer to the no-merging reference at both budgets. These amplitude diagnostics do not establish conservation or full physical consistency. Appendix~\ref{app:weather_diagnostics} provides the full results and definitions.

\section{Discussion}
\label{sec:discussion}

The component and spatial controls support joint allocation by atmospheric structure and input reliability. Dynamics identifies gradients and anomalies, while uncertainty modulates confidence in the input estimate. In the joint score, strong dynamics can counterbalance elevated spread; anchor protection then retains high-scoring structures. The gains in both high-dynamics regimes, together with losses in low-dynamics regions (Table~\ref{tab:regimes_main}), are consistent with concentrating representation capacity on dynamically active regions. This trade-off makes uncertainty most useful when interpreted alongside the atmospheric state.

The choice among E/P/S depends on access to aligned uncertainty products. E uses spread directly, P transfers an externally supervised scoring rule to deployment, and S learns from an input-derived proxy. Their comparison should account for the delayed merge schedule in P/S as well as the final token budget. Within each regime, the reduction ratio should reflect acceptable errors in the target variables: the small-reduction frontier supports accuracy-sensitive use, while larger reductions offer greater encoder savings at a measurable forecast cost.

\section{Conclusion}
\label{sec:conclusion}

We introduced \method, a weather-specific token-allocation policy combining dynamics saliency, relative input uncertainty, anchors, and importance-calibrated matching. Across two ERA5 resolutions, it improves Overall error over ToMe at matched token budgets; controls support contributions from aligned uncertainty and both allocation mechanisms. The three variants accommodate different availability of external uncertainty. Small token reductions preserve most full-token skill, while aggressive compression incurs substantial error. These results support joint dynamics--reliability scoring as an empirical strategy for efficient weather transformers.

\paragraph{Limitations.}
EDA spread estimates relative reanalysis uncertainty and may miss shared model biases or fine-scale extremes. High uncertainty can coincide with strong dynamics; the measured allocation gains coexist with errors in low-dynamics regions. The variance decomposition does not justify a general rule for compressing aleatoric or epistemic uncertainty, for which separate token-level estimators are not evaluated. Experiments cover two transformer backbones and direct 72-hour ERA5 forecasts, without establishing transfer to other architectures or long rollouts. Encoder FLOPs do not measure end-to-end training time. Appendix~\ref{app:limitations} details these boundaries.

\clearpage
\begin{ack}
This manuscript has been authored by UTBattelle, LLC, under contract DE-AC05-00OR22725 with the US Department of Energy (DOE). This research was supported by the ORNL's AI Initiative sponsored by the Director's Research and Development Program at ORNL. This work's computing time was supported through the Oak Ridge Leadership Computing Facility.

\end{ack}
\bibliographystyle{plainnat}
\bibliography{reference}

@article{hersbach2020era5,
  title={The ERA5 global reanalysis},
  author={Hersbach, Hans and Bell, Bill and Berrisford, Paul and Hirahara, Shoji and Hor{\'a}nyi, Andr{\'a}s and Mu{\~n}oz-Sabater, Joaqu{\'\i}n and Nicolas, Julien and Peubey, Carole and Radu, Raluca and Schepers, Dinand and others},
  journal={Quarterly journal of the royal meteorological society},
  volume={146},
  number={730},
  pages={1999--2049},
  year={2020},
  publisher={Wiley Online Library}
}

@article{bolya2022token,
  title={Token merging: Your vit but faster},
  author={Bolya, Daniel and Fu, Cheng-Yang and Dai, Xiaoliang and Zhang, Peizhao and Feichtenhofer, Christoph and Hoffman, Judy},
  journal={arXiv preprint arXiv:2210.09461},
  year={2022}
}

@article{nguyen2023climax,
  title={Climax: A foundation model for weather and climate},
  author={Nguyen, Tung and Brandstetter, Johannes and Kapoor, Ashish and Gupta, Jayesh K and Grover, Aditya},
  journal={arXiv preprint arXiv:2301.10343},
  year={2023}
}

@article{rao2021dynamicvit,
  title={Dynamicvit: Efficient vision transformers with dynamic token sparsification},
  author={Rao, Yongming and Zhao, Wenliang and Liu, Benlin and Lu, Jiwen and Zhou, Jie and Hsieh, Cho-Jui},
  journal={Advances in neural information processing systems},
  volume={34},
  pages={13937--13949},
  year={2021}
}

@inproceedings{KendallGal2017,
 author = {Kendall, Alex and Gal, Yarin},
 title = {What Uncertainties Do We Need in {B}ayesian Deep Learning for Computer Vision?},
 booktitle = {Advances in Neural Information Processing Systems},
 volume = {30},
 year = {2017}
}

@article{HullermeierWaegeman2021,
 author = {H{\"u}llermeier, Eyke and Waegeman, Willem},
 title = {Aleatoric and Epistemic Uncertainty in Machine Learning: An Introduction to Concepts and Methods},
 journal = {Machine Learning},
 volume = {110},
 number = {3},
 pages = {457--506},
 year = {2021},
 doi = {10.1007/s10994-021-05946-3}
}

@inproceedings{Depeweg2018,
 author = {Depeweg, Stefan and Hernandez-Lobato, Jose-Miguel and Doshi-Velez, Finale and Udluft, Steffen},
 title = {Decomposition of Uncertainty in {B}ayesian Deep Learning for Efficient and Risk-sensitive Learning},
 booktitle = {Proceedings of the 35th International Conference on Machine Learning (ICML)},
 year = {2018}
}

@inproceedings{Lakshminarayanan2017,
 author = {Lakshminarayanan, Balaji and Pritzel, Alexander and Blundell, Charles},
 title = {Simple and Scalable Predictive Uncertainty Estimation using Deep Ensembles},
 booktitle = {Advances in Neural Information Processing Systems},
 volume = {30},
 year = {2017}
}

@inproceedings{GalGhahramani2016,
 author = {Gal, Yarin and Ghahramani, Zoubin},
 title = {Dropout as a {B}ayesian Approximation: Representing Model Uncertainty in Deep Learning},
 booktitle = {Proceedings of the 33rd International Conference on Machine Learning (ICML)},
 year = {2016}
}

@inproceedings{ValdenegroToro2024,
 author = {Valdenegro-Toro, Matias and de Jong, Ivo Pascal and Zullich, Marco},
 title = {Unified Uncertainties: Combining Input, Data and Model Uncertainty into a Single Formulation},
 booktitle = {Proceedings of the 41st International Conference on Machine Learning (ICML)},
 year = {2024}
}

@article{MartinElster2022,
 author = {Martin, J{\"o}rg and Elster, Clemens},
 title = {Aleatoric Uncertainty for Errors-in-Variables Models in Deep Regression},
 journal = {Neural Processing Letters},
 year = {2022},
 doi = {10.1007/s11063-022-11066-3}
}

@inproceedings{GastRoth2018,
 author = {Gast, Jochen and Roth, Stefan},
 title = {Lightweight Probabilistic Deep Networks},
 booktitle = {Proceedings of the IEEE Conference on Computer Vision and Pattern Recognition (CVPR)},
 pages = {3369--3378},
 year = {2018}
}

@inproceedings{McHutchonRasmussen2011,
 author = {McHutchon, Andrew and Rasmussen, Carl Edward},
 title = {Gaussian Process Training with Input Noise},
 booktitle = {Advances in Neural Information Processing Systems},
 volume = {24},
 year = {2011}
}

@inproceedings{LiuEtAl2020SNGP,
 author = {Liu, Jeremiah and Lin, Zi and Padhy, Shreyas and Tran, Dustin and Bedrax-Weiss, Tania and Lakshminarayanan, Balaji},
 title = {Simple and Principled Uncertainty Estimation with Deterministic Deep Learning via Distance Awareness},
 booktitle = {Advances in Neural Information Processing Systems},
 volume = {33},
 year = {2020}
}

@inproceedings{Sensoy2018,
 author = {Sensoy, Murat and Kaplan, Lance and Kandemir, Melih},
 title = {Evidential Deep Learning to Quantify Classification Uncertainty},
 booktitle = {Advances in Neural Information Processing Systems},
 volume = {31},
 year = {2018}
}

@article{Janjic2018,
 author = {Janji{\'c}, Tijana and Bormann, Niels and Bocquet, Marc and Carton, James A. and Cohn, Stephen E. and Dance, Sarah L. and Losa, Svetlana N. and Nichols, Nancy K. and Potthast, Roland and Waller, Joanne A. and Weston, Peter},
 title = {On the Representation Error in Data Assimilation},
 journal = {Quarterly Journal of the Royal Meteorological Society},
 volume = {144},
 number = {713},
 pages = {1257--1278},
 year = {2018},
 doi = {10.1002/qj.3130}
}

@book{Fuller1987,
 author = {Fuller, Wayne A.},
 title = {Measurement Error Models},
 publisher = {John Wiley \& Sons},
 year = {1987}
}

@article{Carroll2006,
 author = {Carroll, Raymond J. and Ruppert, David and Stefanski, Leonard A. and Crainiceanu, Ciprian M.},
 title = {Measurement Error in Nonlinear Models: A Modern Perspective},
 journal = {Chapman and Hall/CRC},
 year = {2006},
 edition = {2nd}
}

@article{bi2023accurate,
  title={Accurate medium-range global weather forecasting with 3D neural networks},
  author={Bi, Kaifeng and Xie, Lingxi and Zhang, Hengheng and Chen, Xin and Gu, Xiaotao and Tian, Qi},
  journal={Nature},
  volume={619},
  number={7970},
  pages={533--538},
  year={2023},
  publisher={Nature Publishing Group UK London}
}

@article{lam2023learning,
  title={Learning skillful medium-range global weather forecasting},
  author={Lam, Remi and Sanchez-Gonzalez, Alvaro and Willson, Matthew and Wirnsberger, Peter and Fortunato, Meire and Alet, Ferran and Ravuri, Suman and Ewalds, Timo and Eaton-Rosen, Zach and Hu, Weihua and others},
  journal={Science},
  volume={382},
  number={6677},
  pages={1416--1421},
  year={2023},
  publisher={American Association for the Advancement of Science}
}

@article{tay2022efficient,
  title={Efficient transformers: A survey},
  author={Tay, Yi and Dehghani, Mostafa and Bahri, Dara and Metzler, Donald},
  journal={ACM Computing Surveys},
  volume={55},
  number={6},
  pages={1--28},
  year={2022},
  publisher={ACM New York, NY}
}

@inproceedings{bolya2023token,
  title={Token merging for fast stable diffusion},
  author={Bolya, Daniel and Hoffman, Judy},
  booktitle={Proceedings of the IEEE/CVF conference on computer vision and pattern recognition},
  pages={4599--4603},
  year={2023}
}

@article{zheng2023fast,
  title={Fast training of diffusion models with masked transformers},
  author={Zheng, Hongkai and Nie, Weili and Vahdat, Arash and Anandkumar, Anima},
  journal={arXiv preprint arXiv:2306.09305},
  year={2023}
}

@inproceedings{sehwag2025stretching,
  title={Stretching each dollar: Diffusion training from scratch on a micro-budget},
  author={Sehwag, Vikash and Kong, Xianghao and Li, Jingtao and Spranger, Michael and Lyu, Lingjuan},
  booktitle={Proceedings of the IEEE/CVF Conference on Computer Vision and Pattern Recognition},
  pages={28596--28608},
  year={2025}
}

@inproceedings{krause2025tread,
  title={Tread: Token routing for efficient architecture-agnostic diffusion training},
  author={Krause, Felix and Phan, Timy and Gui, Ming and Baumann, Stefan Andreas and Hu, Vincent Tao and Ommer, Bj{\"o}rn},
  booktitle={Proceedings of the IEEE/CVF International Conference on Computer Vision},
  pages={15703--15713},
  year={2025}
}

@article{park2025sprint,
  title={Sprint: Sparse-Dense Residual Fusion for Efficient Diffusion Transformers},
  author={Park, Dogyun and Haji-Ali, Moayed and Li, Yanyu and Menapace, Willi and Tulyakov, Sergey and Kim, Hyunwoo J and Siarohin, Aliaksandr and Kag, Anil},
  journal={arXiv preprint arXiv:2510.21986},
  year={2025}
}

@inproceedings{wang2025orbit,
  title={ORBIT-2: Scaling Exascale Vision Foundation Models for Weather and Climate Downscaling},
  author={Wang, Xiao and Choi, Jong-Youl and Kurihaya, Takuya and Lyngaas, Isaac and Yoon, Hong-Jun and Xiao, Xi and Pugmire, David and Fan, Ming and Nafi, Nasik Muhammad and Tsaris, Aristeidis and others},
  booktitle={Proceedings of the International Conference for High Performance Computing, Networking, Storage and Analysis},
  pages={86--98},
  year={2025}
}

@misc{chen2025comeconfidenceguidedtokenmerging,
      title={Co-Me: Confidence-Guided Token Merging for Visual Geometric Transformers}, 
      author={Yutian Chen and Yuheng Qiu and Ruogu Li and Ali Agha and Shayegan Omidshafiei and Jay Patrikar and Sebastian Scherer},
      year={2025},
      eprint={2511.14751},
      archivePrefix={arXiv},
      primaryClass={cs.CV},
      url={https://arxiv.org/abs/2511.14751}, 
}

@article{pathak2022fourcastnet,
  title={Fourcastnet: A global data-driven high-resolution weather model using adaptive fourier neural operators},
  author={Pathak, Jaideep and Subramanian, Shashank and Harrington, Peter and Raja, Sanjeev and Chattopadhyay, Ashesh and Mardani, Morteza and Kurth, Thorsten and Hall, David and Li, Zongyi and Azizzadenesheli, Kamyar and others},
  journal={arXiv preprint arXiv:2202.11214},
  year={2022}
}

@article{wang2026global,
  title={Global Attention with Linear Complexity for Exascale Generative Data Assimilation in Earth System Prediction},
  author={Wang, Xiao and Zhang, Zezhong and Lyngaas, Isaac and Yoon, Hong-Jun and Choi, Jong-Youl and Liang, Siming and Wang, Janet and Chipilski, Hristo G and Aji, Ashwin M and Bao, Feng and others},
  journal={arXiv preprint arXiv:2604.16590},
  year={2026}
}

@article{munoz2021era5,
  title={ERA5-Land: A state-of-the-art global reanalysis dataset for land applications},
  author={Mu{\~n}oz-Sabater, Joaqu{\'\i}n and Dutra, Emanuel and Agust{\'\i}-Panareda, Anna and Albergel, Cl{\'e}ment and Arduini, Gabriele and Balsamo, Gianpaolo and Boussetta, Souhail and Choulga, Margarita and Harrigan, Shaun and Hersbach, Hans and others},
  journal={Earth system science data},
  volume={13},
  number={9},
  pages={4349--4383},
  year={2021},
  publisher={Copernicus GmbH}
}

@article{wang2026less,
  title={Less Tokens, Better Forecasts: Sparse Residual Routing for Efficient Weather Prediction},
  author={Wang, Janet and Zhang, Yunbei and Zhao, Lin and Xiao, Xi and Hamm, Jihun and Wang, Xiao},
  journal={arXiv preprint arXiv:2607.02829},
  year={2026}
}

@article{li2026rethinking,
  title={Rethinking layer-wise information allocation for vision foundation model adaptation},
  author={Li, Yuqi and Xiao, Xi and Zhang, Yunbei and Zhao, Lin and Li, Yu and Zhao, Aiden and Wang, Tianyang and Xu, Hao and Tian, Yingli},
  journal={arXiv preprint arXiv:2607.21973},
  year={2026}
}

@inproceedings{zhang2025dpcore,
  title={{DPCore}: Dynamic Prompt Coreset for Continual Test-Time Adaptation},
  author={Zhang, Yunbei and Mehra, Akshay and Niu, Shuaicheng and Hamm, Jihun},
  booktitle={International Conference on Machine Learning},
  pages={75757--75778},
  year={2025},
  organization={PMLR}
}

@article{zhang2026adapting,
  title={Adapting in the Dark: Efficient and Stable Test-Time Adaptation for Black-Box Models},
  author={Zhang, Yunbei and Niu, Shuaicheng and Cai, Chengyi and Liu, Feng and Hamm, Jihun},
  journal={arXiv preprint arXiv:2604.15609},
  year={2026}
}

@inproceedings{zhang2026prime,
  title={Prime Once, then Reprogram Locally: An Efficient Alternative to Black-Box Service Model Adaptation},
  author={Zhang, Yunbei and Cai, Chengyi and Liu, Feng and Hamm, Jihun},
  booktitle={Proceedings of the IEEE/CVF Conference on Computer Vision and Pattern Recognition},
  pages={6176--6187},
  year={2026}
}

@article{xu2026rivatfuse,
  title={{RiVaT-Fuse}: Reliability-Calibrated Variational Tensor Fusion for Multimodal Prediction under Modality Uncertainty},
  author={Xu, Yingfan and Liu, Tieming and Liang, Ye and Liu, Taiping},
  journal={arXiv preprint arXiv:2609.10798},
  year={2026},
  eprint={2609.10798},
  archivePrefix={arXiv},
  url={https://arxiv.org/abs/2609.10798}
}

@article{ouyang2026predicting,
  title={Predicting lasso peptide structure with {LassoPred}},
  author={Ouyang, Xingyu and Ran, Xinchun and Zhu, Dantong and Yang, Zhongyue John},
  journal={Methods in Enzymology},
  volume={730},
  pages={91--106},
  year={2026},
  doi={10.1016/bs.mie.2026.01.037}
}

@article{wu2026quarterly,
  title={Quarterly Signal Persistence Scoring and Temporal Taxonomy for {FAERS} Screening},
  author={Wu, Jialun and Zhang, Peng and Zhu, Dantong and Yang, Tianze and Li, Mingyu},
  journal={SSRN Electronic Journal},
  year={2026},
  month={sep},
  doi={10.2139/ssrn.7404502}
}

@article{zhu2026stress,
  title={Stress-Testing Structure-Aware Calibration of Malware Graph Neural Networks under Type Shift},
  author={Zhu, Junru and Yang, Yixin and Ding, Xiaoqing and Qi, Ruoyu},
  journal={arXiv preprint arXiv:2609.28517},
  year={2026},
  url={https://arxiv.org/abs/2609.28517}
}

@article{zhu2026residual,
  title={Residual Community Prototypes Under-Reject Held-Out Malware Families in {FCG-MFD}},
  author={Zhu, Junru and Yang, Yixin and Ding, Xiaoqing and Qi, Ruoyu},
  journal={arXiv preprint arXiv:2609.24980},
  year={2026},
  url={https://arxiv.org/abs/2609.24980}
}

@inproceedings{zhang2025ot,
  title={{OT-VP}: Optimal Transport-Guided Visual Prompting for Test-Time Adaptation},
  author={Zhang, Yunbei and Mehra, Akshay and Hamm, Jihun},
  booktitle={2025 IEEE/CVF Winter Conference on Applications of Computer Vision (WACV)},
  pages={1122--1132},
  year={2025},
  organization={IEEE},
  url={https://arxiv.org/abs/2407.09498}
}

@article{jin2026deepjepa,
  title={{DeepJEPA}: Scaling World Models from Within},
  author={Jin, Zijian and Zhang, Yunbei and Liu, Yuanzhe and Liu, Ming and Chen, Baian and Ye, Weirui and Liu, Shilong and Pavone, Marco},
  journal={arXiv preprint arXiv:2610.00368},
  year={2026},
  url={https://arxiv.org/abs/2610.00368}
}

@inproceedings{liu2026wan,
  title={{Wan-R1}: Verifiable-Reinforcement Learning for Video Reasoning},
  author={Liu, Ming and Zhang, Yunbei and Liu, Shilong and Wang, Liwen and Zhang, Wensheng},
  booktitle={European Conference on Computer Vision (ECCV)},
  pages={419--436},
  year={2026},
  publisher={Springer},
  doi={10.1007/978-3-032-37461-5_23}
}

@article{zhang2026can,
  title={When Can Text Replace Vision? Structural Bottlenecks in Diagram Reasoning},
  author={Zhang, Yunbei and Wang, Janet and Hamm, Jihun and Reddy, Chandan K},
  journal={arXiv preprint arXiv:2609.39142},
  year={2026},
  url={https://arxiv.org/abs/2609.39142}
}

@article{wang2026medical,
  title={How Medical {VLMs} Underutilize Their Vision Encoders: A Dermatology Perspective},
  author={Wang, Janet and Zhang, Yunbei and Wang, Xiao and Hamm, Jihun},
  journal={arXiv preprint arXiv:2609.36557},
  year={2026},
  url={https://arxiv.org/abs/2609.36557}
}

@article{xi2026prompt,
  title={Prompt-based Adaptation in Large-scale Vision Models: A Survey},
  author={Xiao, Xi and Zhang, Yunbei and Zhao, Lin and Liu, Yiyang and Liao, Xiaoying and Mai, Zheda and Li, Xingjian and Wang, Xiao and Xu, Hao and Hamm, Jihun and Lin, Xue and Xu, Min and Wang, Qifan and Wang, Tianyang and Han, Cheng},
  journal={Transactions on Machine Learning Research},
  year={2026},
  url={https://openreview.net/forum?id=UwtXDttgsE}
}

@article{yu2026elsa3d,
  title={{ELSA3D}: Elastic Semantic Anchoring for Unified {3D} Understanding and Generation},
  author={Yu, Tianjiao and Li, Xinzhuo and Shen, Yifan and Susladkar, Onkar and Liu, Yuanzhe and Zhou, Xiaona and Lourentzou, Ismini},
  journal={arXiv preprint arXiv:2607.06565},
  year={2026}
}
\clearpage
\appendix
\begin{center}
\huge\textbf{Appendix}
\end{center}

\section{Proof of the Three-Way Variance Decomposition}
\label{app:proof}

We give the short derivation used in Sec.~\ref{sec:analytical}. All expectations and variances are conditioned on $\tilde{\bm{x}}$ and $\mathcal{D}$ unless stated otherwise, with the factorized posterior assumed in Sec.~\ref{sec:analytical}. The conditional second moments are finite. Applying the law of total variance over $\bm{x}^*\sim p(\bm{x}^*\mid\tilde{\bm{x}})$ gives
\begin{equation}
    \Var[y]
    =
    \E_{\bm{x}^*}\!\left[\Var(y\mid \bm{x}^*)\right]
    +
    \Var_{\bm{x}^*}\!\left(\E[y\mid \bm{x}^*]\right).
\end{equation}
The second term is the observation uncertainty term $\sigma^2_{\mathrm{obs}}$. For a fixed $\bm{x}^*$, applying the law of total variance again over $\bm{\theta}\sim p(\bm{\theta}\mid\mathcal{D})$ gives
\begin{equation}
    \Var(y\mid \bm{x}^*)
    =
    \E_{\bm{\theta}}\!\left[\Var(y\mid \bm{x}^*,\bm{\theta})\right]
    +
    \Var_{\bm{\theta}}\!\left(\E[y\mid \bm{x}^*,\bm{\theta}]\right).
\end{equation}
Substitution yields Eq.~\eqref{eq:three-way}. If the observation noise is zero, $p(\bm{x}^*\mid\tilde{\bm{x}})$ collapses to a point mass at $\tilde{\bm{x}}$, the observation term vanishes, and the usual aleatoric and epistemic decomposition is recovered.

The delta-method estimate follows by expanding $\mu(\bm{x})=\E_{\bm{\theta}}[f_{\bm{\theta}}(\bm{x})]$ around $\tilde{\bm{x}}$:
\begin{equation}
    \mu(\bm{x}^*) \approx
    \mu(\tilde{\bm{x}})
    +
    \bm{J}_{\mu}(\tilde{\bm{x}})^\top(\bm{x}^*-\tilde{\bm{x}}).
\end{equation}
Taking variance over $\bm{x}^*$ gives
\begin{equation}
\label{eq:delta}
    \sigma^2_{\mathrm{obs}}(\tilde{\bm{x}})
    =
    \Var_{\bm{x}^*}\!\left[\mu(\bm{x}^*)\right]
    \approx
    \bm{J}_{\mu}(\tilde{\bm{x}})^\top
    \bm{\Sigma}_{\bm{x}^*\mid\tilde{\bm{x}}}
    \bm{J}_{\mu}(\tilde{\bm{x}}).
\end{equation}
Here $\bm{\Sigma}_{\bm{x}^*\mid\tilde{\bm{x}}}=\operatorname{Cov}(\bm{x}^*\mid\tilde{\bm{x}})$ is the latent-input posterior covariance. Identifying it with an additive observation-noise covariance requires further assumptions, such as a locally flat latent-state prior and input-independent Gaussian noise. With a diagonal posterior covariance, the first-order term sums squared sensitivities weighted by latent-input variances. This motivates considering both sensitivity and reliability, but does not derive $s_i=d_i-\beta q_i$, an optimal merging rule, or a forecast-error bound. We do not estimate the decomposition's components or use it as a training loss.

\section{Implementation and Evaluation Details}
\label{app:impl}

\textbf{Data splits and tokenization.} The $5.625^\circ$ experiments use the split-sharded ERA5 setting with 2010 to 2014 for training, 2015 for validation, and 2016 for testing. The files are split into 320 train shards, 64 validation shards, and 64 test shards so that file-level DDP sharding remains balanced over the 64-GPU setting. Each variable has spatial shape $32\times64$, and the patch size is $2$, producing a $16\times32$ token grid. The $1.40625^\circ$ experiments use 1979 to 2018 for training, 2019 for validation, and 2020 for testing. Each variable has spatial shape $128\times256$, and the patch size is $4$, producing a $32\times64$ token grid. In both settings, the forecast pair is constructed as input $x_t$ and target $y_{t+72}$.

\textbf{Variables.} The 41 input channels are the static fields land-sea mask, orography, and latitude; the surface variables T2m, U10, and V10; and geopotential, zonal wind, meridional wind, temperature, and specific humidity at pressure levels 50, 250, 500, 600, 700, 850, and 925 hPa. The five forecast targets are Z500, T850, T2m, U10, and V10. Inputs and outputs are normalized using the training-set mean and standard deviation. RMSE and ACC are computed after denormalization.

\textbf{Uncertainty fields.} External UQ denotes ECMWF EDA spread fields aligned with the same time and grid as the ERA5 samples. In the reported runs, UQ is constructed for the full input-channel set used by the forecaster, with the same year split and shard layout as the weather fields. UQ fields are not used as extra forecast input channels. They are normalized per variable, patch-averaged, and used as teacher or calibration signals for token merging. Static variables receive zero observation uncertainty when constructing the full input-channel UQ tensor.

\textbf{Importance score.} Dynamics saliency uses the input-time fields of the selected forecast variables and Eq.~\eqref{eq:dynamics}. Adjacent-grid differences are backward differences, with a zero difference at the first row/column. The anomaly term is centered on the spatial mean of each input channel and has weight $1/2$. The dynamics map and the UQ map are patch-averaged with the model patch size and normalized per sample, yielding $d_i$ and $q_i$. The reported configurations use $s_i=d_i-\beta q_i$ with $\beta=1.0$. This makes high-dynamics, low-spread patches strong anchor candidates and treats high-spread patches as less reliable merge candidates unless the dynamics signal is large. The anchor fraction is $\rho=0.25$. The matching weights in Eq.~\eqref{eq:match} are $\lambda_s=0.05$ and $\lambda_{\Delta}=0.05$ unless otherwise stated.

\textbf{Variants and predictor.} \method-E uses EDA-derived teacher importance directly for merge bias and anchor protection and requires external UQ at evaluation. \method-P uses EDA-derived teacher importance during training, trains a convolutional importance head from the hidden tokens at probe layer 3 (zero-based indexing), and uses predicted importance for merging after a warmup period and during testing. \method-S uses no external UQ: it builds a proxy UQ target from local $3\times3$ spatial inconsistency, trains the same predictor, and uses predicted proxy importance at inference. Predictor-based variants use a pairwise-ranking importance loss with weight $0.2$.

\textbf{Merge schedules.} The main tables use final token-reduction ratios of $25\%$ and $50\%$. At $5.625^\circ$, these remove 128 and 256 tokens from an initial sequence of 512. At $1.40625^\circ$, they remove 512 and 1024 tokens from an initial sequence of 2048. For ToMe and \method-E, ratio-based merging is spread over all eight transformer layers. For \method-P and \method-S, the predictor is produced at probe layer 3, so the current implementation delays ratio-based merging until after the probe layer; the first four layers therefore run with the full token count.

\textbf{Optimization.} Training uses DDP on an HPC cluster of 8 nodes, each exposing $8$ AMD MI250X GPU devices, for $64$ GPU devices in total. We use AdamW with $\beta=(0.9,0.99)$, weight decay $10^{-5}$, peak learning rate $5\times10^{-4}$, 1000 warmup steps, minimum learning rate $10^{-8}$, gradient clipping at $1.0$, periodic validation, and a learning-rate schedule calibrated to 12000 steps. The per-process batch size is 16 at $5.625^\circ$ and 1 at $1.40625^\circ$, corresponding to global batch sizes 1024 and 64 under the 64-GPU setting.

\subsection{Normalization, selection, and evaluation}
For a patch vector $v$, per-sample standardization is
\begin{equation}
\mathcal{Z}(v)_i=\frac{v_i-\bar v}{\max(\operatorname{std}(v),10^{-6})}.
\end{equation}
Thus $d_i=\mathcal{Z}(\operatorname{PatchAvg}(D))_i$ and $q_i$ is the corresponding standardized spread score. A spatially constant spread gives $q_i=0$, recovering the dynamics-only score. Anchor selection protects the highest-scoring fraction when merging begins; protected tokens are excluded from both endpoints of subsequent merges.

We select $\beta$, the anchor fraction, and the matching weights using the held-out validation split, then fix the configuration before evaluating the test set. The sensitivity tables report test performance of the stated settings; the default is $\beta=1$, $\rho=0.25$, and $\lambda_s=\lambda_\Delta=0.05$. The main resolution tables report three-seed means and standard deviations. The additional control tables report point estimates.

For a target field, latitude weighting uses $w_h=\cos(\varphi_h)$ normalized over the evaluated grid. Per-variable wRMSE is the square root of the weighted squared prediction error; Overall averages the five variable-specific errors as in Eq.~\eqref{eq:overall}. ACC is the weighted correlation of predicted and reference anomalies after removal of climatology. We report per-variable errors because their units and magnitudes differ. Supplemental masked diagnostics use per-sample weighted errors and average the valid sample values.

\section{Component Ablations and Parameter Sensitivity}
\label{app:ablations}
The component study compares the full score and allocation mechanism with controlled removals at the same token budgets. Table~\ref{tab:components_full} expands Table~\ref{tab:components_main}. UQ alone is weaker than the combined dynamics--UQ score, and removing either anchors or matching penalties degrades the full method. The anchor removal has the largest effect at $50\%$ reduction.

\begin{table}[!htbp]
\centering
\small
\setlength{\tabcolsep}{7pt}
\caption{\textbf{Both score inputs and both allocation mechanisms contribute.} Overall wRMSE ($\downarrow$) at $1.40625^\circ$. The no-anchor and no-matching rows retain the combined score. Disabling both allocation mechanisms recovers ToMe.}
\label{tab:components_full}
\begin{tabular}{@{}lrr@{}}
\toprule
Configuration & 25\% & 50\% \\
\midrule
    ToMe (neither mechanism) & 42.45 & 67.32 \\
    Dynamics only ($s=d$) & 40.95 & 58.60 \\
    UQ only ($s=-q$) & 41.40 & 61.70 \\
    Combined score, no anchors & 40.31 & 61.40 \\
    Anchors, no matching penalties & 40.68 & 55.90 \\
    \midrule
    \textbf{Full \method-E} & \textbf{39.59} & \textbf{53.00} \\
\bottomrule
\end{tabular}
\end{table}

Table~\ref{tab:sensitivity} varies each major parameter around the validation-selected configuration. Increasing $\beta$ to 4 raises the $50\%$ error to 69.40, and increasing the matching weights to 0.50 raises it to 63.70. Excessive reliance on UQ or matching penalties can therefore overconstrain allocation. Increasing the anchor fraction above 0.25 also raises error. The moderate settings provide the best results in these reported sweeps.

\begin{table}[!htbp]
\centering
\small
\setlength{\tabcolsep}{7pt}
\caption{\textbf{Moderate parameter values balance protection and compression.} One-parameter sweeps at $1.40625^\circ$, with other settings held at the default. Columns report Overall wRMSE ($\downarrow$). The validation-selected default values are $\beta=1$, $\rho=0.25$, and $\lambda_s=\lambda_\Delta=0.05$.}
\label{tab:sensitivity}
\begin{tabular}{@{}lrrr@{}}
\toprule
Parameter & Value & 25\% & 50\% \\
\midrule
    $\beta$ & 0 & 40.95 & 58.60 \\
     & 0.25 & 40.12 & 55.10 \\
     & 0.5 & 39.74 & 53.62 \\
     & 1 & 39.59 & 53.00 \\
     & 2 & 40.36 & 54.85 \\
     & 4 & 43.80 & 69.40 \\
    \midrule
    $\rho$ & 0 & 40.31 & 61.40 \\
     & 0.05 & 40.05 & 58.20 \\
     & 0.10 & 39.78 & 55.30 \\
     & 0.25 & 39.59 & 53.00 \\
     & 0.40 & 39.86 & 53.94 \\
     & 0.50 & 40.62 & 56.80 \\
    \midrule
    $\lambda_s=\lambda_\Delta$ & 0 & 40.68 & 55.90 \\
     & 0.01 & 40.02 & 54.20 \\
     & 0.05 & 39.59 & 53.00 \\
     & 0.10 & 39.71 & 53.35 \\
     & 0.20 & 40.94 & 56.10 \\
     & 0.50 & 44.60 & 63.70 \\
\bottomrule
\end{tabular}
\end{table}
\section{Robustness to Uncertainty Estimates}
\label{app:uq_robustness}
\subsection{Spatial controls and monotone transformations}
The UQ controls preserve the weather input and token budget while modifying the uncertainty field. Spatial shuffling retains the spread values but changes their locations. The random-noise control matches the per-channel scale statistics of the reference spread. Constant UQ removes the standardized uncertainty contribution. Square-root and square transformations change magnitude spacing, and rank mapping replaces spread magnitudes with normalized spatial ranks. The sign control changes only the score direction to $s=d+\beta q$.

Table~\ref{tab:uq_robust_full} reports both forecast performance and agreement with the original importance signal. Rank correlation compares patch-score orderings; anchor overlap measures the shared fraction of protected tokens. The combined score includes dynamics as well as UQ, so its ordering can change under a monotone transform of UQ. The measured stability is an empirical property of these transformations.
\begin{table}[!htbp]
\centering
\footnotesize
\setlength{\tabcolsep}{6pt}
\caption{\textbf{Tested rescalings preserve useful spatial order.} Score rank correlation and anchor overlap are relative to the original EDA field. The last two columns are Overall wRMSE ($\downarrow$) at $1.40625^\circ$. Dashes mark agreement metrics not reported for the constant field.}
\label{tab:uq_robust_full}
\begin{tabular}{@{}lrrrr@{}}
\toprule
UQ field & Rank corr. & Anchor overlap & 25\% & 50\% \\
\midrule
    Original EDA & 1.000 & 100\% & 39.59 & 53.00 \\
    $\sqrt{\mathrm{UQ}}$ & 0.995 & 94\% & 39.62 & 53.05 \\
    $\mathrm{UQ}^2$ & 0.977 & 88\% & 39.68 & 53.19 \\
    Rank-mapped UQ & 0.969 & 84\% & 39.72 & 53.28 \\
    Simulated 2-member & 0.960 & 79\% & 39.94 & 54.10 \\
    Constant UQ & -- & -- & 40.95 & 58.60 \\
    Shuffled UQ & $\approx0$ & 26\% & 41.88 & 62.30 \\
\bottomrule
\end{tabular}
\end{table}
\begin{table}[!htbp]
\centering
\small
\setlength{\tabcolsep}{7pt}
\caption{\textbf{Removing spatial information or reversing the UQ direction degrades performance.} Overall wRMSE ($\downarrow$) at $1.40625^\circ$.}
\label{tab:uq_controls_full}
\begin{tabular}{@{}lrr@{}}
\toprule
Control & 25\% & 50\% \\
\midrule
    Original EDA spread & 39.59 & 53.00 \\
    Constant UQ / dynamics only & 40.95 & 58.60 \\
    Spatially shuffled UQ & 41.88 & 62.30 \\
    Matched random-noise UQ & 42.07 & 63.15 \\
    Inverted sign ($s=d+\beta q$) & 42.64 & 64.90 \\
\bottomrule
\end{tabular}
\end{table}

\subsection{Simulated ensemble sampling noise}
To study finite-ensemble estimation noise, we perturb the reference EDA spread at several nominal ensemble-size settings. ERA5 EDA itself uses ten members; the counts in this simulation label estimator-noise settings, rather than observed subsets or additional EDA members. Table~\ref{tab:ensemble_members} reports score correlations and forecast errors. These results use simulated spread perturbations, not subsampling of archived raw ensemble members.
\begin{table}[!htbp]
\centering
\small
\setlength{\tabcolsep}{7pt}
\caption{\textbf{Finite-ensemble sampling noise causes gradual degradation.} This experiment simulates noisier spread estimates, rather than subsampling archived raw ensemble members. Rank correlation is measured against the reference importance score; errors are Overall wRMSE at $1.40625^\circ$.}
\label{tab:ensemble_members}
\begin{tabular}{@{}lrrr@{}}
\toprule
Simulated members & Rank corr. & 25\% & 50\% \\
\midrule
    Full reference & 1.000 & 39.59 & 53.00 \\
    32 & 0.999 & 39.60 & 53.02 \\
    16 & 0.997 & 39.62 & 53.08 \\
    8 & 0.994 & 39.66 & 53.21 \\
    4 & 0.986 & 39.76 & 53.55 \\
    2 & 0.960 & 39.94 & 54.10 \\
\bottomrule
\end{tabular}
\end{table}
\section{Weather-Regime, Extreme-Value, and Structural Diagnostics}
\label{app:weather_diagnostics}
\subsection{Regime definitions and aggregation}
Table~\ref{tab:regimes_main} uses four masks defined by the per-sample medians of the patch-level UQ and dynamics fields. A patch is assigned to the high group when its value is at least the corresponding median. Patch masks are expanded to their spatial support. For a mask $A$ and one forecast sample, we compute
\begin{equation}
\mathrm{wRMSE}(A)=\sqrt{\frac{\sum_{h,w}w_h\,\mathbf{1}_A(h,w)(\hat y_{h,w}-y_{h,w})^2}
{\sum_{h,w}w_h\,\mathbf{1}_A(h,w)}}.
\end{equation}
Valid per-sample errors are averaged over the evaluation samples. Each quadrant is normalized by its own weight sum; the reported Global error is evaluated over the full field, not an unweighted average of the four quadrant errors. At $25\%$ reduction, the high-dynamics improvements of 34.6 and 48.4 Z500 wRMSE coexist with low-dynamics increases of 17.8 and 25.3, respectively. This describes where the budget reallocation helps and where it incurs a cost.

\subsection{Tail errors}
For each sample and target variable, upper/lower-tail masks select values above/below the spatial 95th/5th percentiles of the reference anomaly field. We evaluate prediction error in the target's physical units within those masks using the weighted expression above. Table~\ref{tab:tail_errors} shows the complete upper/lower Z500 and upper T2m results.
\begin{table}[!htbp]
\centering
\footnotesize
\setlength{\tabcolsep}{7pt}
\caption{\textbf{\method-E better retains forecast extremes than ToMe at both budgets.} Conditional wRMSE ($\downarrow$) on the upper/lower $5\%$ target-anomaly masks at $1.40625^\circ$. Z500 errors use $\mathrm{m^2\,s^{-2}}$ and T2m errors use K.}
\label{tab:tail_errors}
\begin{tabular}{@{}lrrrr@{}}
\toprule
Method & Reduction & Z500 upper & Z500 lower & T2m upper \\
\midrule
    No merging & 0\% & 189.9 & 242.5 & 1.953 \\
    \midrule
    ToMe & 25\% & 251.4 & 318.7 & 2.318 \\
    \method-E & 25\% & 226.8 & 285.3 & 2.147 \\
    \midrule
    ToMe & 50\% & 402.6 & 512.4 & 3.145 \\
    \method-E & 50\% & 312.5 & 396.1 & 2.612 \\
\bottomrule
\end{tabular}
\end{table}

\subsection{Spatial and wind-structure ratios}
The standard-deviation ratio is the predicted spatial standard deviation divided by the reference standard deviation. The gradient ratio compares latitude-weighted mean gradient magnitudes, using physical grid spacing. For U10/V10, kinetic energy per unit mass is $K=(u^2+v^2)/2$, and the relative-vorticity diagnostic uses the finite-difference expression $\zeta=\partial_xv-\partial_yu$. The KE ratio compares weighted mean $K$ and the vorticity ratio compares weighted mean $|\zeta|$ between prediction and reference. Ratios are formed per sample and then averaged. Longitude differences are periodic for these diagnostics, and latitude differences use one-sided boundary handling. A ratio of one indicates matching amplitude statistics.

Table~\ref{tab:structure_ratios} shows improved preservation of spatial variability, gradients, and wind statistics relative to ToMe.
\begin{table}[!htbp]
\centering
\footnotesize
\setlength{\tabcolsep}{5pt}
\caption{\textbf{Spatial and wind statistics remain closer to the full-token reference.} Prediction/reference ratios at $1.40625^\circ$; unity denotes agreement with the reference statistic. The spatial standard-deviation and gradient columns concern Z500; KE and vorticity use U10/V10.}
\label{tab:structure_ratios}
\begin{tabular}{@{}lrrrrr@{}}
\toprule
Method & Reduction & $\sigma$ ratio & $\nabla$ ratio & KE ratio & Vort. ratio \\
\midrule
    No merging & 0\% & 0.998 & 1.013 & 0.975 & 0.870 \\
    \midrule
    ToMe & 25\% & 0.964 & 1.092 & 0.931 & 0.812 \\
    \method-E & 25\% & 0.981 & 1.048 & 0.958 & 0.846 \\
    \midrule
    ToMe & 50\% & 0.907 & 1.213 & 0.874 & 0.731 \\
    \method-E & 50\% & 0.955 & 1.104 & 0.936 & 0.802 \\
\bottomrule
\end{tabular}
\end{table}
\section{Additional Baselines and Backbone}
\label{app:baselines}
\subsection{Pruning controls}
Table~\ref{tab:pruning_baselines} compares random, dynamics-only, UQ-only, and joint-score selection with merging. The comparison holds the token budget and schedule fixed across policies. The joint score improves over the strongest alternative pruning score by 2.64 at $25\%$ reduction and 2.56 at $50\%$. Holding that joint score fixed, \method-E merging further improves error by 1.15 and 4.21, respectively. Both the allocation signal and the representation-combining operator matter in this setting.
\begin{table}[!htbp]
\centering
\small
\setlength{\tabcolsep}{7pt}
\caption{\textbf{The joint score benefits pruning, and merging preserves more forecast skill.} Overall wRMSE ($\downarrow$) at $1.40625^\circ$, using the matched backbone, split, schedule, and token budgets of the comparison. Encoder FLOPs are $0.86\times$/$0.55\times$ at $25\%$/$50\%$ reduction.}
\label{tab:pruning_baselines}
\begin{tabular}{@{}lrr@{}}
\toprule
Method & 25\% & 50\% \\
\midrule
    Random pruning & 61.86 & 85.90 \\
    Dynamics-saliency pruning & 44.42 & 61.02 \\
    UQ-only pruning & 43.38 & 59.77 \\
    \method-score pruning & 40.74 & 57.21 \\
    \midrule
    ToMe merging & 42.45 & 67.32 \\
    \textbf{\method-E merging} & \textbf{39.59} & \textbf{53.00} \\
\bottomrule
\end{tabular}
\end{table}

\subsection{A plain 12-layer ViT}
The second backbone uses a single joint patch embedding, twelve transformer layers, and no per-variable embeddings or variable-aggregation attention. The same score, anchor protection, and matching mechanism apply to its spatial tokens. Table~\ref{tab:plain_vit} reports matched-FLOPs comparisons within this backbone. At $25\%$ reduction, \method-E obtains 141.13 versus 145.05 for ToMe; at $50\%$, it obtains 149.44 versus 157.14. These results extend the evidence beyond ClimaX-style variable tokenization to another patch-transformer architecture.
\begin{table}[!htbp]
\centering
\small
\setlength{\tabcolsep}{7pt}
\caption{\textbf{The allocation rule transfers to a plain patch-transformer backbone.} Overall wRMSE ($\downarrow$) on a 12-layer ViT with a joint patch embedding. FLOPs are relative to this backbone without merging.}
\label{tab:plain_vit}
\begin{tabular}{@{}lrrr@{}}
\toprule
Method & Reduction & wRMSE & Encoder FLOPs \\
\midrule
    No merging & 0\% & 141.96 & $1.000\times$ \\
    \midrule
    ToMe & 25\% & 145.05 & $0.829\times$ \\
    \method-E & 25\% & 141.13 & $0.829\times$ \\
    \midrule
    ToMe & 50\% & 157.14 & $0.661\times$ \\
    \method-E & 50\% & 149.44 & $0.661\times$ \\
\bottomrule
\end{tabular}
\end{table}
\section{Predictor Quality and Merge-Decision Analysis}
\label{app:predictor_quality}

Table~\ref{tab:predictor_quality} reports the held-out patch-level predictor alignment summarized in Sec.~\ref{sec:predictor}. The EDA-derived teacher and the self-supervised proxy define different prediction targets; correlations should be read within each target. Forecast performance in the main tables provides the downstream measure of allocation quality.

\begin{table}[h]
    \centering
    \small
    \setlength{\tabcolsep}{5pt}
    \caption{Predictor quality on held-out $5.625^\circ$ cases. The case label denotes the spatial resolution, not a separate dataset. Correlation and z-normalized MAE are computed patch by patch between the predictor output and its supervision target.}
    \label{tab:predictor_quality}
    \begin{tabular}{@{}lcccc@{}}
        \toprule
        Case & \method-P corr. $\uparrow$ & \method-P zMAE $\downarrow$ & \method-S corr. $\uparrow$ & \method-S zMAE $\downarrow$ \\
        \midrule
        $5.625^\circ$ case 01 & $0.820$ & $0.466$ & $0.858$ & $0.403$ \\
        $5.625^\circ$ case 02 & $0.825$ & $0.463$ & $0.900$ & $0.341$ \\
        $5.625^\circ$ case 03 & $0.769$ & $0.525$ & $0.870$ & $0.396$ \\
        \bottomrule
    \end{tabular}
\end{table}

\subsection{Merge-set and importance statistics}
\label{app:merge_decisions}
Let $E_T$ and $E_U$ denote the selected merge-edge sets for ToMe and \method. Their Jaccard overlap, $|E_T\cap E_U|/|E_T\cup E_U|$, is 0.139 at $25\%$ reduction. Table~\ref{tab:merge_statistics} summarizes the accompanying importance diagnostics. The first two rows concern the merged-token statistics; the last row concerns the displayed pair endpoints in Fig.~\ref{fig:cosine_failure}. These use different aggregation and score-display conventions and are reported separately. The figure's top-$15\%$ circle markers are a visualization aid, whereas algorithmic protection uses the $25\%$ anchor fraction and the allocation diagnostic uses the top importance quartile.
\begin{table}[!htbp]
\centering
\footnotesize
\setlength{\tabcolsep}{7pt}
\caption{\textbf{\method\ changes the selected merge set.} Allocation diagnostics at $25\%$ token reduction. Quartile membership and merged-token importance refer to the standardized importance score.}
\label{tab:merge_statistics}
\begin{tabular}{@{}lrr@{}}
\toprule
Diagnostic & ToMe & \method \\
\midrule
    Merged tokens in top quartile & 46.8\% & 0\% \\
    Mean merged-token importance & +0.732 & -1.456 \\
    Displayed-pair mean (Fig.~\ref{fig:cosine_failure}) & 0.28 & 0.11 \\
\bottomrule
\end{tabular}
\end{table}
\section{Additional Qualitative Visualizations}
\label{app:additional_qual}

\begin{figure}[!t]
    \centering
    \begin{subfigure}[t]{0.455\linewidth}
        \centering
        \includegraphics[width=\linewidth]{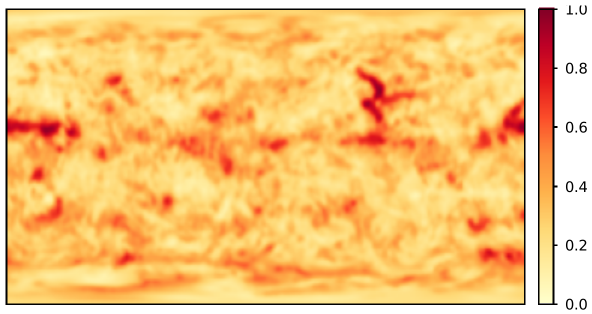}
        \caption{\textbf{Combined observation uncertainty.}}
    \end{subfigure}
    \hfill
    \begin{subfigure}[t]{0.455\linewidth}
        \centering
        \includegraphics[width=\linewidth]{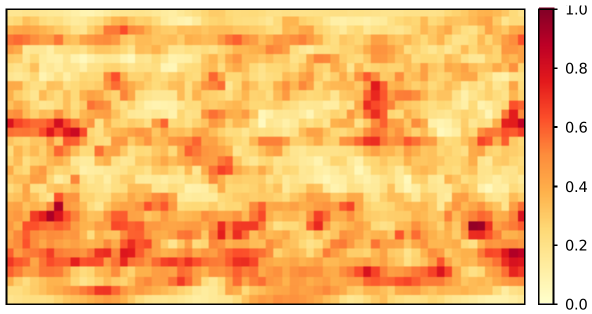}
        \caption{\textbf{Patch-level token score.}}
    \end{subfigure}
    \vspace{-1.5mm}
    \caption{\textbf{Motivation for uncertainty-guided token merging at $1.40625^\circ$.} The observation-uncertainty field is not uniform noise; it contains coherent spatial structure. After patch aggregation and combination with dynamics saliency, the token-score grid separates low-uncertainty high-saliency regions from high-spread merge candidates.}
    \label{fig:uq_to_token}
    \vspace{-2mm}
\end{figure}
\begin{figure}[h]
    \centering
    \includegraphics[width=0.95\linewidth]{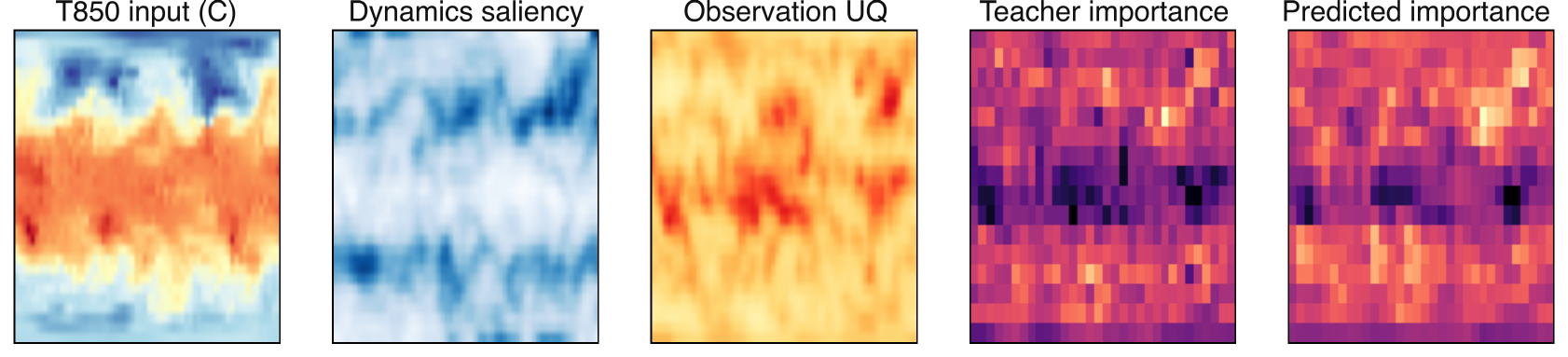}
    \caption{\textbf{Additional importance-score decomposition.} From left to right, the figure shows the input weather state, dynamics saliency, observation uncertainty, teacher importance, and predicted importance for a held-out $5.625^\circ$ case.}
    \label{fig:app_importance_breakdown}
\end{figure}

Fig.~\ref{fig:app_importance_breakdown} expands the score construction for a held-out $5.625^\circ$ case. The dynamics map highlights fronts and high-gradient structures, while the observation-UQ map highlights regions with larger EDA spread. The teacher and predicted importance maps are not simple copies of either input map; they preserve broad dynamically important bands while suppressing less reliable high-spread regions. This supports the design choice of combining dynamics and uncertainty before deciding which tokens to protect.

\begin{figure}[h]
    \centering
    \includegraphics[width=0.68\linewidth]{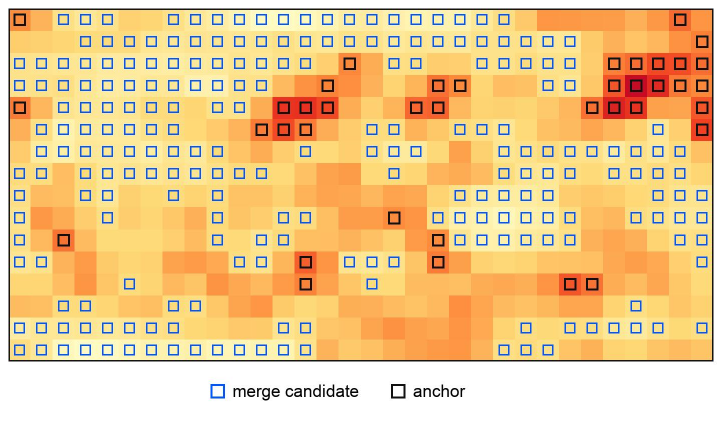}
    \caption{\textbf{Additional anchor and merge-candidate overlay at $5.625^\circ$.} Blue squares denote merge candidates and black squares denote anchors selected from the patch-level token score.}
    \label{fig:app_anchor_overlay}
\end{figure}

Fig.~\ref{fig:app_anchor_overlay} shows the final token-level decision induced by the score. Anchor tokens concentrate in high-importance regions that should retain separate representations, while merge candidates occupy lower-score regions where local redundancy is higher. The visualization complements Fig.~\ref{fig:cosine_failure}: the former shows the absolute anchor and candidate layout, and the latter shows how this layout changes the selected merge edges relative to cosine-only matching.

\begin{figure}[h]
    \centering
    \includegraphics[width=0.90\linewidth]{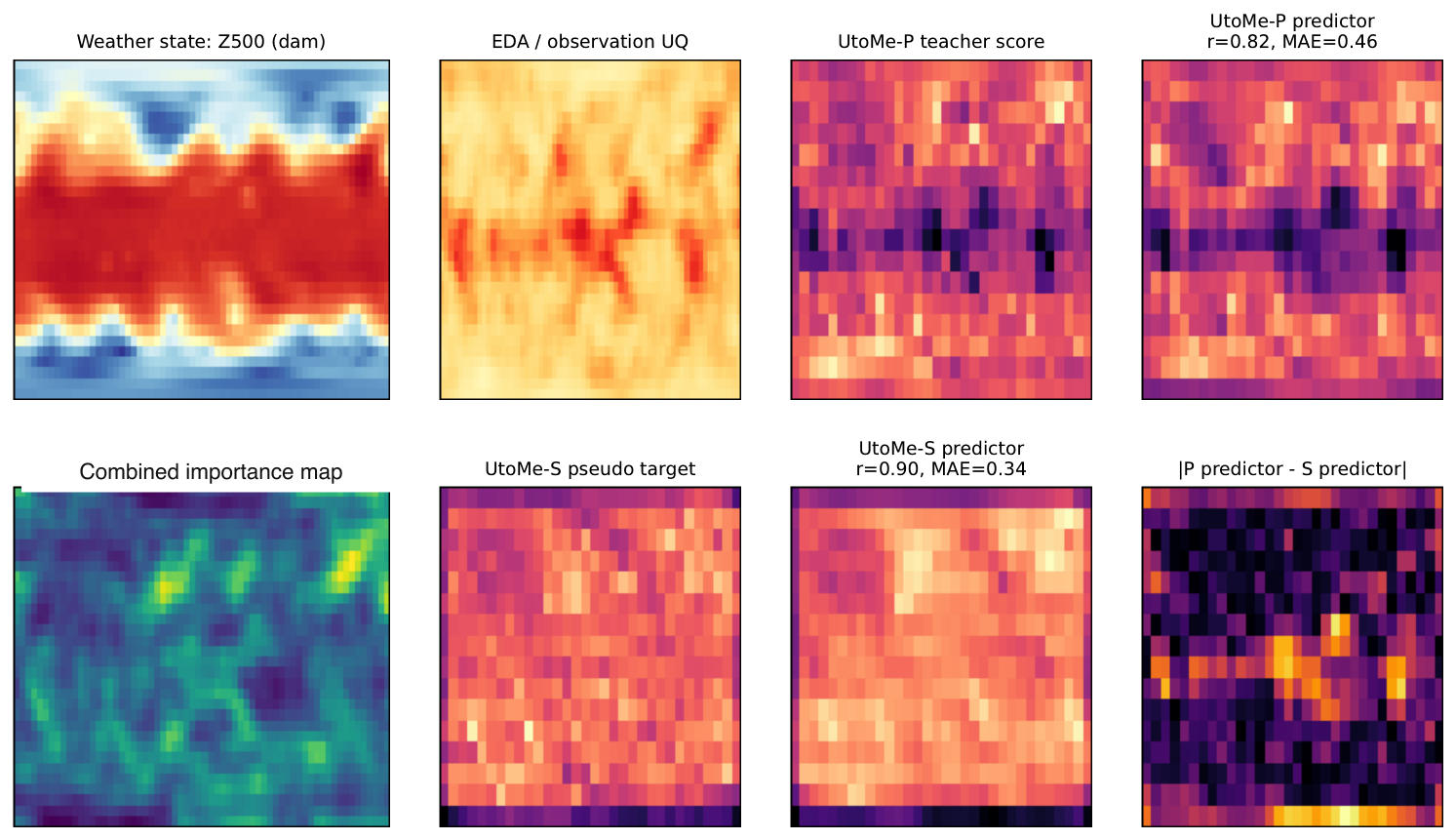}
    \caption{\textbf{Predictor visualization for \method-P and \method-S.} The learned predictors recover the broad spatial structure of their supervision targets, which explains why the predictor-based variants remain close to \method-E in the main results.}
    \label{fig:app_predictor_comparison}
\end{figure}

Fig.~\ref{fig:app_predictor_comparison} visualizes the predictor setting used by \method-P and \method-S. The top row compares the EDA-derived teacher score with the \method-P predictor, while the bottom row compares the self-supervised pseudo target with the \method-S predictor. The visual agreement is not perfect, but the broad high- and low-score regions are preserved. This agreement supports learning a useful score for allocation. Anchor selection uses score ordering, while pair penalties additionally depend on score magnitudes.

\begin{figure}[h]
    \centering
    \begin{subfigure}[t]{0.47\linewidth}
        \centering
        \includegraphics[width=\linewidth]{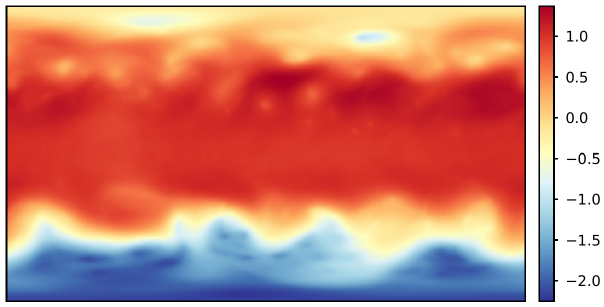}
        \caption{\textbf{Z500 input.}}
    \end{subfigure}
    \hfill
    \begin{subfigure}[t]{0.47\linewidth}
        \centering
        \includegraphics[width=\linewidth]{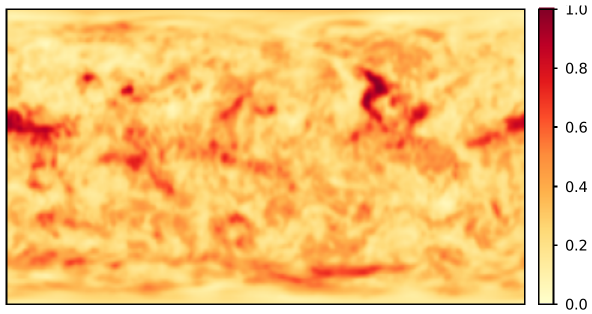}
        \caption{\textbf{Observation uncertainty.}}
    \end{subfigure}

    \vspace{0.5em}
    \begin{subfigure}[t]{0.47\linewidth}
        \centering
        \includegraphics[width=\linewidth]{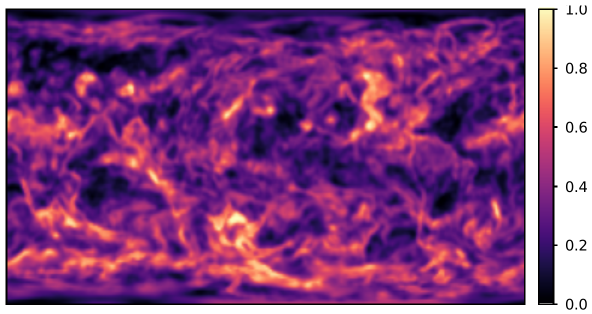}
        \caption{\textbf{Importance score.}}
    \end{subfigure}
    \hfill
    \begin{subfigure}[t]{0.47\linewidth}
        \centering
        \includegraphics[width=\linewidth]{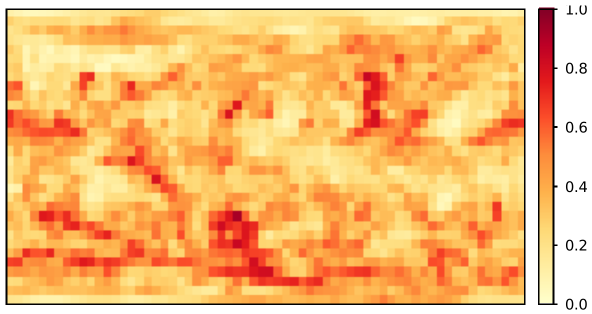}
        \caption{\textbf{Patch-level token score.}}
    \end{subfigure}
    \caption{\textbf{Additional $1.40625^\circ$ case study.} The panels show how a high-resolution weather state and its aligned observation-uncertainty field are converted into the importance and token-score maps used by \method.}
    \label{fig:app_highres_case}
\end{figure}

Fig.~\ref{fig:app_highres_case} provides an additional high-resolution case study at $1.40625^\circ$. The observation-uncertainty field contains coherent bands and localized high-spread structures that do not simply duplicate the weather state. After combining this field with dynamics saliency, the patch-level token score becomes smoother and more directly usable by the merge policy. This case supports the same qualitative conclusion as the main visualizations: uncertainty-guided scoring gives the model a structured basis for deciding which regions should be protected and which regions can be merged.

\section{Limitations}
\label{app:limitations}

\textbf{Uncertainty semantics and calibration.} EDA spread reflects uncertainty in the assimilated reanalysis state, including observation and model effects, rather than isolating sensor noise. ECMWF describes it as a relative uncertainty estimate; common biases and under-resolved cyclone structure can produce small spread despite substantial error. Spatial and temporal alignment, per-variable spread scaling, and changes in the observing system can affect the score. Our monotone-transformation and simulated sampling-noise controls do not establish robustness to arbitrary distribution shift. The simulated ensemble-size experiment does not subsample raw EDA members.

\textbf{Allocation and difficult weather.} Dynamics saliency is a heuristic based on gradients and anomalies, not a calibrated forecast sensitivity. High uncertainty need not imply irrelevance: fronts, data-sparse regions, and rapidly developing events can be both uncertain and important. The regime analysis supports joint scoring on the tested data while recording losses in low-dynamics regions. Conditional tail errors are not event-specific cyclone or precipitation evaluations. Gradient and wind-amplitude ratios are useful structural diagnostics but do not verify conservation laws or complete physical consistency.

\textbf{Analytical and architectural scope.} The variance identity is conditional on the stated posterior factorization and does not prove the direction or optimality of the allocation policy. EDA does not supply separate calibrated token-level aleatoric and epistemic estimates, so their effects on merging are not directly compared. We evaluate a ClimaX-style backbone and a plain ViT, not GraphCast, FourCastNet, Pangu-Weather, or other tokenizations. The evidence concerns direct 72-hour forecasts at two ERA5 resolutions; long autoregressive rollouts, operational forecasting, and other modalities remain untested.

\textbf{Metrics, compute, and reproducibility.} Overall averages errors with different physical units and is dominated by Z500; the per-variable metrics should guide interpretation. At high resolution, $25\%$ and $50\%$ token reductions incur material losses relative to no merging. Encoder FLOPs exclude data processing, uncertainty estimation, and other end-to-end costs, and do not establish proportional wall-clock savings. P/S have different merging schedules and slightly higher compute than E/ToMe. Additional controls are point estimates rather than three-seed comparisons. Exact reproducibility additionally requires the final experiment code, preprocessing configuration, and evaluation scripts.
\newpage
\section*{NeurIPS Paper Checklist}

\begin{enumerate}

\item {\bf Claims}
    \item[] Question: Do the main claims made in the abstract and introduction accurately reflect the paper's contributions and scope?
    \item[] Answer: \answerYes{}
    \item[] Justification: The abstract and Sec.~\ref{sec:intro} state the contributions of \method\ as observation-uncertainty-guided token merging with three UQ-availability variants, and these claims are supported by Sec.~\ref{sec:method} and the experiments in Sec.~\ref{sec:experiments}.
    \item[] Guidelines:
    \begin{itemize}
        \item The answer \answerNA{} means that the abstract and introduction do not include the claims made in the paper.
        \item The abstract and/or introduction should clearly state the claims made, including the contributions made in the paper and important assumptions and limitations. A \answerNo{} or \answerNA{} answer to this question will not be perceived well by the reviewers. 
        \item The claims made should match theoretical and experimental results, and reflect how much the results can be expected to generalize to other settings. 
        \item It is fine to include aspirational goals as motivation as long as it is clear that these goals are not attained by the paper. 
    \end{itemize}

\item {\bf Limitations}
    \item[] Question: Does the paper discuss the limitations of the work performed by the authors?
    \item[] Answer: \answerYes{}
    \item[] Justification: Appendix~\ref{app:limitations} discusses the main limitations: experiments cover ERA5 forecasting at two resolutions on the primary backbone and an additional plain ViT comparison; the discussion also covers UQ shift, finite-ensemble simulation, difficult regimes, and the distinction between encoder FLOPs and end-to-end efficiency.
    \item[] Guidelines:
    \begin{itemize}
        \item The answer \answerNA{} means that the paper has no limitation while the answer \answerNo{} means that the paper has limitations, but those are not discussed in the paper. 
        \item The authors are encouraged to create a separate ``Limitations'' section in their paper.
        \item The paper should point out any strong assumptions and how robust the results are to violations of these assumptions (e.g., independence assumptions, noiseless settings, model well-specification, asymptotic approximations only holding locally). The authors should reflect on how these assumptions might be violated in practice and what the implications would be.
        \item The authors should reflect on the scope of the claims made, e.g., if the approach was only tested on a few datasets or with a few runs. In general, empirical results often depend on implicit assumptions, which should be articulated.
        \item The authors should reflect on the factors that influence the performance of the approach. For example, a facial recognition algorithm may perform poorly when image resolution is low or images are taken in low lighting. Or a speech-to-text system might not be used reliably to provide closed captions for online lectures because it fails to handle technical jargon.
        \item The authors should discuss the computational efficiency of the proposed algorithms and how they scale with dataset size.
        \item If applicable, the authors should discuss possible limitations of their approach to address problems of privacy and fairness.
        \item While the authors might fear that complete honesty about limitations might be used by reviewers as grounds for rejection, a worse outcome might be that reviewers discover limitations that aren't acknowledged in the paper. The authors should use their best judgment and recognize that individual actions in favor of transparency play an important role in developing norms that preserve the integrity of the community. Reviewers will be specifically instructed to not penalize honesty concerning limitations.
    \end{itemize}

\item {\bf Theory assumptions and proofs}
    \item[] Question: For each theoretical result, does the paper provide the full set of assumptions and a complete (and correct) proof?
    \item[] Answer: \answerYes{}
    \item[] Justification: The three-way variance decomposition is stated in Sec.~\ref{sec:analytical} with finite second moments and an explicit factorized latent-input/parameter posterior; Appendix~\ref{app:proof} gives the full derivation and the delta-method estimate.
    \item[] Guidelines:
    \begin{itemize}
        \item The answer \answerNA{} means that the paper does not include theoretical results. 
        \item All the theorems, formulas, and proofs in the paper should be numbered and cross-referenced.
        \item All assumptions should be clearly stated or referenced in the statement of any theorems.
        \item The proofs can either appear in the main paper or the supplemental material, but if they appear in the supplemental material, the authors are encouraged to provide a short proof sketch to provide intuition. 
        \item Inversely, any informal proof provided in the core of the paper should be complemented by formal proofs provided in appendix or supplemental material.
        \item Theorems and Lemmas that the proof relies upon should be properly referenced. 
    \end{itemize}

    \item {\bf Experimental result reproducibility}
    \item[] Question: Does the paper fully disclose all the information needed to reproduce the main experimental results of the paper to the extent that it affects the main claims and/or conclusions of the paper (regardless of whether the code and data are provided or not)?
    \item[] Answer: \answerYes{}
    \item[] Justification: Sec.~\ref{sec:setup} and Appendix~\ref{app:impl} document data splits, variable lists, model architecture, merge schedule, importance-score configuration, and optimizer settings, which together specify the procedure used to produce the reported numbers.
    \item[] Guidelines:
    \begin{itemize}
        \item The answer \answerNA{} means that the paper does not include experiments.
        \item If the paper includes experiments, a \answerNo{} answer to this question will not be perceived well by the reviewers: Making the paper reproducible is important, regardless of whether the code and data are provided or not.
        \item If the contribution is a dataset and\slash or model, the authors should describe the steps taken to make their results reproducible or verifiable. 
        \item Depending on the contribution, reproducibility can be accomplished in various ways. For example, if the contribution is a novel architecture, describing the architecture fully might suffice, or if the contribution is a specific model and empirical evaluation, it may be necessary to either make it possible for others to replicate the model with the same dataset, or provide access to the model. In general. releasing code and data is often one good way to accomplish this, but reproducibility can also be provided via detailed instructions for how to replicate the results, access to a hosted model (e.g., in the case of a large language model), releasing of a model checkpoint, or other means that are appropriate to the research performed.
        \item While NeurIPS does not require releasing code, the conference does require all submissions to provide some reasonable avenue for reproducibility, which may depend on the nature of the contribution. For example
        \begin{enumerate}
            \item If the contribution is primarily a new algorithm, the paper should make it clear how to reproduce that algorithm.
            \item If the contribution is primarily a new model architecture, the paper should describe the architecture clearly and fully.
            \item If the contribution is a new model (e.g., a large language model), then there should either be a way to access this model for reproducing the results or a way to reproduce the model (e.g., with an open-source dataset or instructions for how to construct the dataset).
            \item We recognize that reproducibility may be tricky in some cases, in which case authors are welcome to describe the particular way they provide for reproducibility. In the case of closed-source models, it may be that access to the model is limited in some way (e.g., to registered users), but it should be possible for other researchers to have some path to reproducing or verifying the results.
        \end{enumerate}
    \end{itemize}

\item {\bf Open access to data and code}
    \item[] Question: Does the paper provide open access to the data and code, with sufficient instructions to faithfully reproduce the main experimental results, as described in supplemental material?
    \item[] Answer: \answerYes{}
    \item[] Justification: ERA5 is publicly available. The project page linked in the abstract provides access to the public code repository, including the E/P/S implementations, training and checkpoint evaluation, data-preparation instructions, and configurations for the main forecast comparisons.
    \item[] Guidelines:
    \begin{itemize}
        \item The answer \answerNA{} means that paper does not include experiments requiring code.
        \item Please see the NeurIPS code and data submission guidelines (\url{https://neurips.cc/public/guides/CodeSubmissionPolicy}) for more details.
        \item While we encourage the release of code and data, we understand that this might not be possible, so \answerNo{} is an acceptable answer. Papers cannot be rejected simply for not including code, unless this is central to the contribution (e.g., for a new open-source benchmark).
        \item The instructions should contain the exact command and environment needed to run to reproduce the results. See the NeurIPS code and data submission guidelines (\url{https://neurips.cc/public/guides/CodeSubmissionPolicy}) for more details.
        \item The authors should provide instructions on data access and preparation, including how to access the raw data, preprocessed data, intermediate data, and generated data, etc.
        \item The authors should provide scripts to reproduce all experimental results for the new proposed method and baselines. If only a subset of experiments are reproducible, they should state which ones are omitted from the script and why.
        \item At submission time, to preserve anonymity, the authors should release anonymized versions (if applicable).
        \item Providing as much information as possible in supplemental material (appended to the paper) is recommended, but including URLs to data and code is permitted.
    \end{itemize}

\item {\bf Experimental setting/details}
    \item[] Question: Does the paper specify all the training and test details (e.g., data splits, hyperparameters, how they were chosen, type of optimizer) necessary to understand the results?
    \item[] Answer: \answerYes{}
    \item[] Justification: Sec.~\ref{sec:setup} reports the data splits, model configuration, optimizer, learning-rate schedule, and merge ratios; Appendix~\ref{app:impl} gives the full implementation details.
    \item[] Guidelines:
    \begin{itemize}
        \item The answer \answerNA{} means that the paper does not include experiments.
        \item The experimental setting should be presented in the core of the paper to a level of detail that is necessary to appreciate the results and make sense of them.
        \item The full details can be provided either with the code, in appendix, or as supplemental material.
    \end{itemize}

\item {\bf Experiment statistical significance}
    \item[] Question: Does the paper report error bars suitably and correctly defined or other appropriate information about the statistical significance of the experiments?
    \item[] Answer: \answerYes{}
    \item[] Justification: Tables~\ref{tab:main_5p625} and~\ref{tab:main_1p4} report mean and standard deviation over three random seeds. Additional control tables give point estimates; Appendix~\ref{app:impl} distinguishes these from the repeated-run main results.
    \item[] Guidelines:
    \begin{itemize}
        \item The answer \answerNA{} means that the paper does not include experiments.
        \item The authors should answer \answerYes{} if the results are accompanied by error bars, confidence intervals, or statistical significance tests, at least for the experiments that support the main claims of the paper.
        \item The factors of variability that the error bars are capturing should be clearly stated (for example, train/test split, initialization, random drawing of some parameter, or overall run with given experimental conditions).
        \item The method for calculating the error bars should be explained (closed form formula, call to a library function, bootstrap, etc.)
        \item The assumptions made should be given (e.g., Normally distributed errors).
        \item It should be clear whether the error bar is the standard deviation or the standard error of the mean.
        \item It is OK to report 1-sigma error bars, but one should state it. The authors should preferably report a 2-sigma error bar than state that they have a 96\% CI, if the hypothesis of Normality of errors is not verified.
        \item For asymmetric distributions, the authors should be careful not to show in tables or figures symmetric error bars that would yield results that are out of range (e.g., negative error rates).
        \item If error bars are reported in tables or plots, the authors should explain in the text how they were calculated and reference the corresponding figures or tables in the text.
    \end{itemize}

\item {\bf Experiments compute resources}
    \item[] Question: For each experiment, does the paper provide sufficient information on the computer resources (type of compute workers, memory, time of execution) needed to reproduce the experiments?
    \item[] Answer: \answerNo{}
    \item[] Justification: Sec.~\ref{sec:setup} and Appendix~\ref{app:impl} report the GPU type and count, batch sizes, and training schedule. Per-run execution time, peak memory use, and total project compute are not reported.
    \item[] Guidelines:
    \begin{itemize}
        \item The answer \answerNA{} means that the paper does not include experiments.
        \item The paper should indicate the type of compute workers CPU or GPU, internal cluster, or cloud provider, including relevant memory and storage.
        \item The paper should provide the amount of compute required for each of the individual experimental runs as well as estimate the total compute. 
        \item The paper should disclose whether the full research project required more compute than the experiments reported in the paper (e.g., preliminary or failed experiments that didn't make it into the paper). 
    \end{itemize}
    
\item {\bf Code of ethics}
    \item[] Question: Does the research conducted in the paper conform, in every respect, with the NeurIPS Code of Ethics \url{https://neurips.cc/public/EthicsGuidelines}?
    \item[] Answer: \answerYes{}
    \item[] Justification: The work uses publicly available reanalysis data, does not involve human subjects, and the authors have reviewed and follow the NeurIPS Code of Ethics.
    \item[] Guidelines:
    \begin{itemize}
        \item The answer \answerNA{} means that the authors have not reviewed the NeurIPS Code of Ethics.
        \item If the authors answer \answerNo, they should explain the special circumstances that require a deviation from the Code of Ethics.
        \item The authors should make sure to preserve anonymity (e.g., if there is a special consideration due to laws or regulations in their jurisdiction).
    \end{itemize}

\item {\bf Broader impacts}
    \item[] Question: Does the paper discuss both potential positive societal impacts and negative societal impacts of the work performed?
    \item[] Answer: \answerNA{}
    \item[] Justification: The work is a methodological contribution to efficient training of weather forecasting models on the publicly available ERA5 reanalysis, and we do not identify a direct path to negative societal impact distinct from the broader weather-modeling literature.
    \item[] Guidelines:
    \begin{itemize}
        \item The answer \answerNA{} means that there is no societal impact of the work performed.
        \item If the authors answer \answerNA{} or \answerNo, they should explain why their work has no societal impact or why the paper does not address societal impact.
        \item Examples of negative societal impacts include potential malicious or unintended uses (e.g., disinformation, generating fake profiles, surveillance), fairness considerations (e.g., deployment of technologies that could make decisions that unfairly impact specific groups), privacy considerations, and security considerations.
        \item The conference expects that many papers will be foundational research and not tied to particular applications, let alone deployments. However, if there is a direct path to any negative applications, the authors should point it out. For example, it is legitimate to point out that an improvement in the quality of generative models could be used to generate Deepfakes for disinformation. On the other hand, it is not needed to point out that a generic algorithm for optimizing neural networks could enable people to train models that generate Deepfakes faster.
        \item The authors should consider possible harms that could arise when the technology is being used as intended and functioning correctly, harms that could arise when the technology is being used as intended but gives incorrect results, and harms following from (intentional or unintentional) misuse of the technology.
        \item If there are negative societal impacts, the authors could also discuss possible mitigation strategies (e.g., gated release of models, providing defenses in addition to attacks, mechanisms for monitoring misuse, mechanisms to monitor how a system learns from feedback over time, improving the efficiency and accessibility of ML).
    \end{itemize}
    
\item {\bf Safeguards}
    \item[] Question: Does the paper describe safeguards that have been put in place for responsible release of data or models that have a high risk for misuse (e.g., pre-trained language models, image generators, or scraped datasets)?
    \item[] Answer: \answerNA{}
    \item[] Justification: The paper releases an efficient training method for weather forecasters and does not release pretrained generative models, scraped datasets, or other artifacts with a high risk of misuse.
    \item[] Guidelines:
    \begin{itemize}
        \item The answer \answerNA{} means that the paper poses no such risks.
        \item Released models that have a high risk for misuse or dual-use should be released with necessary safeguards to allow for controlled use of the model, for example by requiring that users adhere to usage guidelines or restrictions to access the model or implementing safety filters. 
        \item Datasets that have been scraped from the Internet could pose safety risks. The authors should describe how they avoided releasing unsafe images.
        \item We recognize that providing effective safeguards is challenging, and many papers do not require this, but we encourage authors to take this into account and make a best faith effort.
    \end{itemize}

\item {\bf Licenses for existing assets}
    \item[] Question: Are the creators or original owners of assets (e.g., code, data, models), used in the paper, properly credited and are the license and terms of use explicitly mentioned and properly respected?
    \item[] Answer: \answerYes{}
    \item[] Justification: ERA5 reanalysis data~\citep{hersbach2020era5} is used under the Copernicus Climate Data Store license, the ClimaX-style model and data conventions are credited to~\citet{nguyen2023climax} (MIT license), and ToMe~\citep{bolya2022token,bolya2023token} is credited as the train-time merging baseline.
    \item[] Guidelines:
    \begin{itemize}
        \item The answer \answerNA{} means that the paper does not use existing assets.
        \item The authors should cite the original paper that produced the code package or dataset.
        \item The authors should state which version of the asset is used and, if possible, include a URL.
        \item The name of the license (e.g., CC-BY 4.0) should be included for each asset.
        \item For scraped data from a particular source (e.g., website), the copyright and terms of service of that source should be provided.
        \item If assets are released, the license, copyright information, and terms of use in the package should be provided. For popular datasets, \url{paperswithcode.com/datasets} has curated licenses for some datasets. Their licensing guide can help determine the license of a dataset.
        \item For existing datasets that are re-packaged, both the original license and the license of the derived asset (if it has changed) should be provided.
        \item If this information is not available online, the authors are encouraged to reach out to the asset's creators.
    \end{itemize}

\item {\bf New assets}
    \item[] Question: Are new assets introduced in the paper well documented and is the documentation provided alongside the assets?
    \item[] Answer: \answerYes{}
    \item[] Justification: The released code includes a README, data-preparation instructions, training and evaluation scripts, configuration files, and licensing information. The paper does not release a new dataset or pretrained model.
    \item[] Guidelines:
    \begin{itemize}
        \item The answer \answerNA{} means that the paper does not release new assets.
        \item Researchers should communicate the details of the dataset\slash code\slash model as part of their submissions via structured templates. This includes details about training, license, limitations, etc. 
        \item The paper should discuss whether and how consent was obtained from people whose asset is used.
        \item At submission time, remember to anonymize your assets (if applicable). You can either create an anonymized URL or include an anonymized zip file.
    \end{itemize}

\item {\bf Crowdsourcing and research with human subjects}
    \item[] Question: For crowdsourcing experiments and research with human subjects, does the paper include the full text of instructions given to participants and screenshots, if applicable, as well as details about compensation (if any)?
    \item[] Answer: \answerNA{}
    \item[] Justification: The work does not involve crowdsourcing or any research with human subjects.
    \item[] Guidelines:
    \begin{itemize}
        \item The answer \answerNA{} means that the paper does not involve crowdsourcing nor research with human subjects.
        \item Including this information in the supplemental material is fine, but if the main contribution of the paper involves human subjects, then as much detail as possible should be included in the main paper. 
        \item According to the NeurIPS Code of Ethics, workers involved in data collection, curation, or other labor should be paid at least the minimum wage in the country of the data collector. 
    \end{itemize}

\item {\bf Institutional review board (IRB) approvals or equivalent for research with human subjects}
    \item[] Question: Does the paper describe potential risks incurred by study participants, whether such risks were disclosed to the subjects, and whether Institutional Review Board (IRB) approvals (or an equivalent approval/review based on the requirements of your country or institution) were obtained?
    \item[] Answer: \answerNA{}
    \item[] Justification: The work does not involve human subjects, so no IRB approval is required.
    \item[] Guidelines:
    \begin{itemize}
        \item The answer \answerNA{} means that the paper does not involve crowdsourcing nor research with human subjects.
        \item Depending on the country in which research is conducted, IRB approval (or equivalent) may be required for any human subjects research. If you obtained IRB approval, you should clearly state this in the paper. 
        \item We recognize that the procedures for this may vary significantly between institutions and locations, and we expect authors to adhere to the NeurIPS Code of Ethics and the guidelines for their institution. 
        \item For initial submissions, do not include any information that would break anonymity (if applicable), such as the institution conducting the review.
    \end{itemize}

\item {\bf Declaration of LLM usage}
    \item[] Question: Does the paper describe the usage of LLMs if it is an important, original, or non-standard component of the core methods in this research? Note that if the LLM is used only for writing, editing, or formatting purposes and does \emph{not} impact the core methodology, scientific rigor, or originality of the research, declaration is not required.
    \item[] Answer: \answerNA{}
    \item[] Justification: LLMs are not used as an important, original, or non-standard component of the proposed method; any incidental use is limited to writing and formatting and therefore does not require declaration.
    \item[] Guidelines:
    \begin{itemize}
        \item The answer \answerNA{} means that the core method development in this research does not involve LLMs as any important, original, or non-standard components.
        \item Please refer to our LLM policy in the NeurIPS handbook for what should or should not be described.
    \end{itemize}

\end{enumerate}

\end{document}